\pdfoutput=1
\documentclass[11pt]{article}

\usepackage[preprint]{acl}

\usepackage{times}
\usepackage{latexsym}

\usepackage[T1]{fontenc}
\usepackage[utf8]{inputenc}

\usepackage{microtype}

\usepackage{inconsolata}

\usepackage{graphicx}
\usepackage{booktabs}
\usepackage{multirow}
\usepackage{hyperref}
\usepackage{fontawesome} 
\usepackage{makecell}
\usepackage{wrapfig}  
\usepackage{adjustbox}
\usepackage{mathrsfs}
\usepackage{listings} 
\usepackage{pifont}
\usepackage{subfigure}
\usepackage{alltt}  
\usepackage{subcaption}
\usepackage{url}
\usepackage{svg}
\usepackage{makecell}
\usepackage{amsmath}
\usepackage{cleveref}
\crefname{section}{§}{§§}
\Crefname{section}{§}{§§}
\usepackage{amssymb}
\usepackage{svg}
\usepackage[normalem]{ulem} 
\usepackage{colortbl}
\usepackage{multirow}
\usepackage{algorithm}
\usepackage{algpseudocode}
\usepackage[figuresright]{rotating}
\usepackage[most]{tcolorbox}
\usepackage{amssymb}  % 用于 \checkmark 和 \texttimes
\usepackage{natbib}   % 用于 \citep
\usepackage{array}
\usepackage{wasysym}
\usepackage{CJKutf8}
\usepackage{caption}
\usepackage{ragged2e}
\usepackage{longtable}
\usepackage{booktabs}
\usepackage{tabularx}
\usepackage{enumitem}

\definecolor{BoxBackground}{RGB}{240, 240, 240} % 浅灰色背景
\definecolor{BoxFrame}{RGB}{0, 0, 0} % 黑色边框
\definecolor{TitleBackground}{RGB}{0, 0, 0} % 标题背景颜色
\definecolor{TitleText}{RGB}{255, 255, 255} % 标题文字颜色
\tcbset{
  academicbox/.style={
    boxsep=5pt,
    left=2pt,
    right=2pt,
    bottom=0.5pt,
    boxrule=0.5pt,
    colback=BoxBackground,
    colframe=BoxFrame,
    colbacktitle=TitleBackground,
    coltitle=TitleText,
    enhanced,
    attach boxed title to top left={yshift=-0.1in,xshift=0.1in},
    boxed title style={boxrule=0pt,colframe=white},
    title={#1},
  }
}
\newtcolorbox{AcademicBox}[1][]{academicbox=#1}

\definecolor{s_doc_qa_c}{HTML}{da0d68}
\definecolor{m_doc_qa_c}{HTML}{da1d23}
\definecolor{summarization_C}{HTML}{ebb40f}
\definecolor{dialogue_c}{HTML}{187a2e}
\definecolor{synthetic_c}{HTML}{0aa3b5}

\title{Benchmarking Chinese Commonsense Reasoning with a Multi-hop Reasoning Perspective}

\author{
Wangjie You$^1{}^2$\thanks{\ \ Contributed equally. Work was done when Wangjie was interning at ByteDance.}  \quad Xusheng Wang$^1$\footnotemark[1] \quad 
Xing Wang$^1$ \quad \\ \textbf{Wenxiang Jiao} \quad \textbf{Chao Feng}$^1$$^\dagger$ \quad \bf Juntao Li$^2$\thanks{Chao Feng and Juntao Li are co‐corresponding authors.} \quad Min Zhang$^2$ \\
$^{1}$Douyin Content Group, ByteDance \\
$^{2}$School of Computer Science and Technology, Soochow University \\
$^1$\texttt{\{youwangjie, wangxusheng.m, chaofeng.zz\}@bytedance.com} \\
$^2$\texttt{\{ljt, minzhang\}@suda.edu.cn} 
}

\begin{document}
\begin{CJK}{UTF8}{gkai}

\maketitle

\begin{abstract}
While Large Language Models (LLMs) have demonstrated advanced reasoning capabilities, their comprehensive evaluation in general Chinese-language contexts remains understudied.
%While Large Language Models (LLMs) have demonstrated impressive reasoning capabilities, comprehensive evaluation of these abilities within the general scenarios of the Chinese language remains a significant gap in current research.
To bridge this gap, we propose \textbf{C}hinese \textbf{C}ommonsense \textbf{M}ulti-h\textbf{O}p \textbf{R}easoning~(CCMOR), a novel benchmark designed to evaluate LLMs' ability to integrate Chinese-specific factual knowledge with multi-step logical reasoning.
%To address the issue, we introduce a novel benchmark Chinese Commonsense Multi-Hop Reasoning （CCMR） for evaluating Chinese multi-hop commonsense reasoning in LLMs. The CCMR assesses LLMs' capacity to leverage Chinese-specific factual recall and apply multi-step reasoning skills. 
Specifically, we first construct a domain-balanced seed set from existing QA datasets, then develop an LLM-powered pipeline to generate multi-hop questions anchored on factual unit chains.
%Specifically, we leverage existing QA datasets as a domain-balanced seed set and implement an LLM-driven expansion pipeline to generate multi-hop questions anchored on factual unit chains. 
To ensure the quality of resulting dataset, we implement a human-in-the-loop verification system, where domain experts systematically validate and refine the generated questions.
%To ensure the quality and validity of the resulting dataset, we integrate a human-in-the-loop approach where expert annotators rigorously verify and refine the generated questions.
Using CCMOR, we evaluate state-of-the-art LLMs, demonstrating persistent limitations in LLMs' ability to process long-tail knowledge and execute knowledge-intensive reasoning. 
Notably, retrieval-augmented generation substantially mitigates these knowledge gaps, yielding significant performance gains.
%Our experiments reveal that LLMs still exhibit weaknesses in handling long-tail factor knowledge and performing knowledge-intensive reasoning tasks. Additionally, augmenting these models with retrieval capabilities effectively addresses the knowledge gap, leading to significant improvements in performance. 
The dataset will be released upon acceptance.
\end{abstract}
\begin{figure*}[t]
    \centering
    \includegraphics[width=0.96\linewidth]{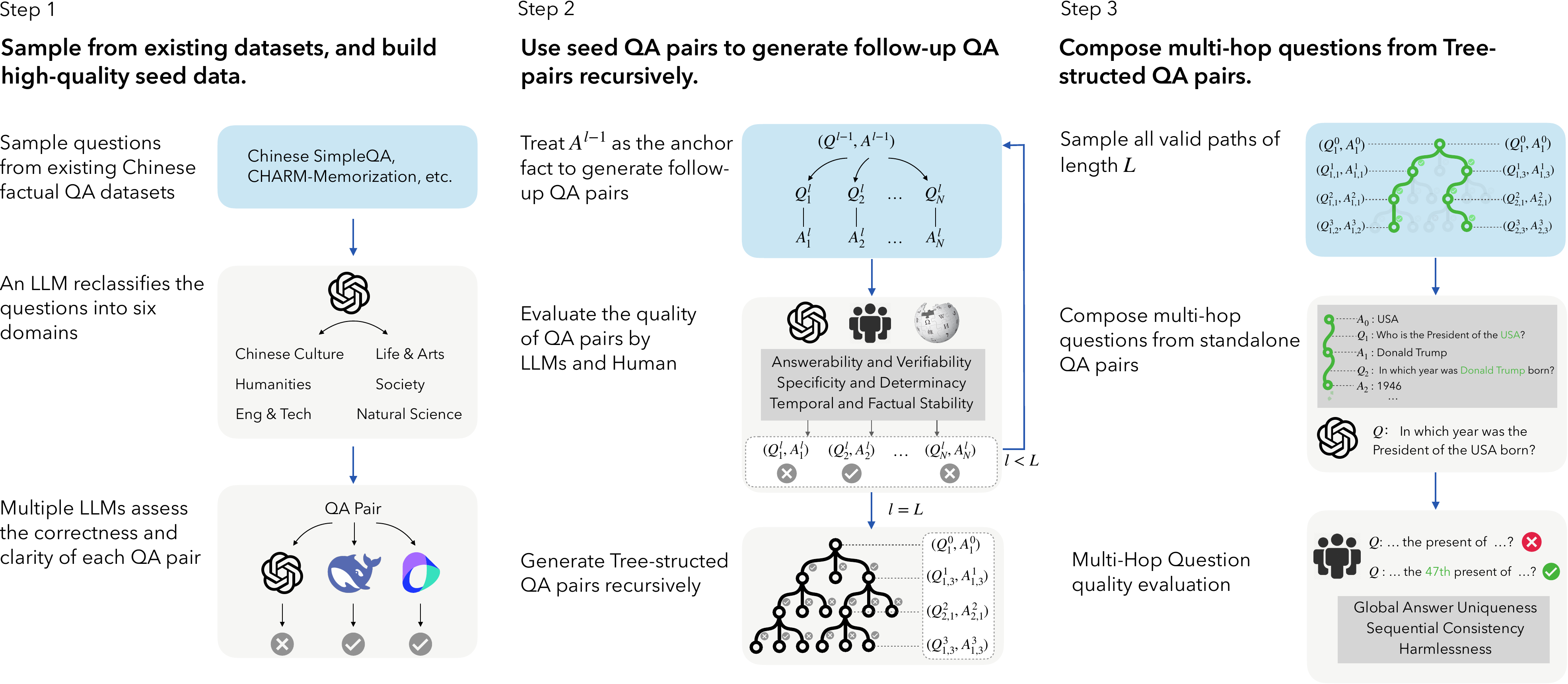}
    \caption{
        An overview of the data construction process. Examples are provided in English for readability.
    }
    \label{fig:pipeline}
    \vspace{-5pt}
\end{figure*}

\section{Introduction}
% math resasoning --> general reasoning
%Large language models (LLMs) have recently received significant attention due to their remarkable reasoning capability, e.g., OpenAI-o1~\cite{jaech2024openai}, DeepSeek-R1~\cite{guo2025deepseek}, Kimi k1.5~\cite{team2025kimi}, Qwen-QwQ~\cite{team2025qwq}. These reasoning-specialized models have shown remarkable advancements in complex tasks across diverse fields such as science, coding, and mathematics. However, despite their remarkable progress, the evaluation of LLMs reasoning capabilities within the general scenarios of the Chinese language domain remains underexplored.

Recent advances in large language models (LLMs) have demonstrated exceptional reasoning capabilities, as exemplified by models like OpenAI-o1~\cite{jaech2024openai}, DeepSeek-R1~\cite{guo2025deepseek}, Kimi k1.5~\cite{team2025kimi}, Qwen-QwQ~\cite{team2025qwq}. These specialized models have achieved significant breakthroughs in complex tasks such as scientific reasoning, programming, mathematical problem-solving and so on. However, their reasoning performance in general Chinese-language scenarios remains understudied, creating a critical research gap.

%general reasoning --> Chinese general reasoning
Multi-hop reasoning, which requires integrating and synthesizing information from multiple sources to conclude, is a crucial aspect of advanced reasoning skills  in the general scenarios ~\cite{welbl2018constructing}. Existing datasets for evaluating multi-hop reasoning, e.g., HotpotQA~\cite{yang2018hotpotqa}, WikiHop~\cite{welbl2018constructing}, DROP~\cite{dua2019drop}, mainly focus on the English language, leaving a significant resource gap for evaluating Chinese LLMs on the reasoning skill.

% Chellenges in Chinese general reasoning
Unfortunately, constructing a high-quality Chinese multi-hop reasoning dataset faces several challenges:  1) \textit{Cultural Relevance}: The dataset needs to be rooted in Chinese cultural knowledge, idioms, and logical reasoning patterns, which differ from the widely used English datasets.  2) \textit{Breadth of Knowledge}: Covering a diverse range of domains within the vast scope of Chinese culture and knowledge is crucial for comprehensively evaluating the reasoning abilities.  3) \textit{Reasoning over Memorization}: The dataset should prioritize tasks requiring reasoning over simple memorization.  4) \textit{Quality Assurance}: Ensuring the accuracy, consistency, and clarity of question-answer pairs demands accurate design and rigorous quality control measures specifically tailored to the nuances of the Chinese language.

To bridge this gap, we propose \textbf{C}hinese \textbf{C}ommonsense \textbf{M}ulti-h\textbf{O}p \textbf{R}easoning~(CCMOR), a novel benchmark designed to evaluate LLMs' ability to integrate Chinese-specific factual knowledge with multi-step logical reasoning.
Partially inspired by MoreHopQA~\cite{schnitzler2024morehopqa}, we leverage existing QA datasets to build a balanced seed set, and then develop an LLM-powered pipeline to generate multi-hop questions anchored on factual unit chains.
To ensure the quality of resulting dataset, we employ human-in-the-loop verification to validate and refine the generated questions.
%To address the above challenges, we propose a novel multi-hop reasoning benchmark for evaluating Chinese commonsense reasoning. 
%Motived by the MoreHopQA~\cite{schnitzler2024morehopqa}, we use the existing multi-hop dataset to synthetic the multi-hop question-answer pairs. To ensure quality and relevance, we leverage the advanced LLMs  to assess each synthetic QA pair sample.   Finally,  we employ human-in-the-loop verification to validate and refine the synthetic data. A case ``港珠澳大桥连接着香港、珠海和一个地区，这个地区曾赴葡萄牙修读语言及法律课程的行政长官，他在20世纪80年代考入了哪所大学？''.

Our contributions are summarized as follows:
\begin{itemize}[leftmargin=15pt]
\item We introduce a novel benchmark CCMOR for evaluating the ability of LLMs in Chinese Commonsense Multi-Hop Reasoning.
%We present  a novel benchmark dataset for evaluating Chinese multi-hop commonsense rea-soning in LLMs.

\item Experimental results on state-of-the-art LLMs demonstrate persistent limitations in LLMs' ability to process long-tail knowledge and execute knowledge-intensive reasoning.
%We evaluated the representative LLMs and find that LLMs still exhibit weaknesses in handling long-tail factor knowledge and performing knowledge-intensive reasoning tasks. 

\item Further analysis suggests that domains requiring procedural or abstract reasoning are more challenging for LLMs, while LLMs with deliberate thinking are more capable of solving multi-hop questions. Notably, retrieval-augmented generation can mitigate the knowledge gaps and improve performance significantly.
\end{itemize}

\section{Chinese Commonsense Multi-Hop Reasoning Dataset}

\begin{table*}[t]
\centering
\small
\begin{tabular}{p{0.45\linewidth} | p{0.45\linewidth}}
\toprule
\textcolor{green}{\smiley} \textbf{Accepted Sub-questions} & \textcolor{blue}{\frownie} \textbf{Rejected Sub-questions} \\
\midrule
\textbf{Q:} 新中国是在哪一年成立的？ \newline
 In which year was the People's Republic of China founded? \newline
\textit{\small  [Concrete and specific]}  
&
\textbf{Q:} 新中国是什么时候成立的？ \newline
 When did the People's Republic of China founded? \newline
\newline
\textit{\small [Vague instruction]}  \\
\midrule
\textbf{Q:} 日本的首都是哪座城市？ \newline
 What is the capital city of Japan? \newline
\textit{\small [Objective and verifiable]}
&
\textbf{Q:} 你认为日本最好的城市是哪一个？ \newline
 Which city in Japan do you think is the best? \newline
\textit{\small [Subjective and unverifiable]} \\
\midrule
\textbf{Q:} 第四十七任届美国总统是谁？ \newline
 Who is the 47th president of the United States? \newline
\textit{\small [Specific and temporally stable]}
&
\textbf{Q:} 现任美国总统是谁？ \newline
 Who is the current president of the United States? \newline
\textit{\small [Temporally unstable and context-dependent]} \\
\bottomrule
\end{tabular}
\caption{Examples of accepted and rejected sub-questions based on our quality criteria.}
\label{tab:subqa_quality_examples_zh_en}
\vspace{-5pt}
\end{table*}

\subsection{Overview}
We introduce the Chinese Commonsense Multi-Hop Reasoning Benchmark~(CCMOR), filling the gap in benchmarks for evaluating multi-step reasoning capabilities of LLMs in Chinese. Our benchmark systematically assesses both factual recall and multi-hop inferential reasoning.
Building upon existing Chinese commonsense QA datasets, we construct a domain-balanced seed set and develop an LLM-driven expansion pipeline to generate multi-hop questions anchored on fact chains. To ensure both factual accuracy and reasoning coherence, the pipeline integrates automatic validation by LLMs with expert human verification.
Figure~\ref{fig:pipeline} illustrates the overall data construction process.
The resulting dataset covers a wide range of domains and features verifiable multi-hop reasoning paths, providing a comprehensive resource to evaluate how LLMs memorize, inference, and reasoning within Chinese commonsense scenarios.

\subsection{Dataset Construction}
\paragraph{Seed Data Sampling}

We sample seed instances from existing Chinese factual QA datasets, including \textit{Chinese SimpleQA}~\cite{he2024chinesesimpleqachinesefactuality}, \textit{CHARM-Memorization}~\cite{sun2024benchmarking}, and others. Following the taxonomy of the Chinese SimpleQA dataset, we prompt LLM to reclassify all seed questions into six primary domains: \textit{Chinese Culture}, \textit{Humanities}, \textit{Engineering and Technology}, \textit{Life and Arts}, \textit{Society}, and \textit{Natural Science}. To ensure domain representativeness, we maintain a balanced distribution across categories during sampling.

To improve data quality and enable multi-hop expansion, we employ multiple LLMs to assess the correctness and clarity of each QA pair. Specifically, the models verify whether a question is logically well-formed and unambiguous, and whether its answer aligns with up-to-date factual knowledge. To support structured multi-hop reasoning, we further ensure that each answer corresponds to a well-defined factual unit (e.g., a person, location, date and so on). This approach facilitates systematic expansion in subsequent stages.

% Through this rigorous selection and validation pipeline, we curate a high-quality seed set of 1,000 instances with balanced coverage across fine-grained domains.

\paragraph{Sub-question Generation}

Given a seed QA pair \((q_1^0, a_1^0)\), we treat the answer \(a_1^0\) as the \textit{anchor fact} for constructing follow-up questions. At each layer \(\ell \in [1, N]\), where \(N\) is a predefined maximum depth, we prompt an LLM to generate \(n\) new QA pairs for each \((q^{\ell-1}_i, a^{\ell-1}_i) \in \mathcal{QA}_{\ell-1}\), based on the anchor fact \(a^{\ell-1}_i\):

\begin{equation}
\mathcal{QA}_\ell = \bigcup_{i \in \mathcal{QA}_{\ell-1}} \{(q^\ell_{i,1}, a^\ell_{i,1}), \dots, (q^\ell_{i,n}, a^\ell_{i,n})\}.
\end{equation}

To promote diversity and reduce model-specific bias, different LLMs are alternated across layers. This recursive expansion yields a tree-structured set \(\{\mathcal{QA}_1, \dots, \mathcal{QA}_N\}\), where each node corresponds to an answer, and each directed edge represents a question that connects an answer to its follow-up. A complete path from the root to a leaf forms a coherent multi-hop question, with each hop grounded in a preceding factual answer.

\paragraph{LLM-Based Verification}

For each generated sub-question, we employ LLMs to assess its quality and determine whether to retain the corresponding node for further expansion.

Unlike traditional QA datasets~\cite{he2024chinesesimpleqachinesefactuality} that assume a single ground-truth answer per question, we do not enforce the uniqueness of the answers at the sub-question level. Instead, the answer of each sub-question is concrete, countable, and verifiable. For sub-questions with multiple valid answers, we retain all plausible answers, but only one is selected for further expansion.
Table~\ref{tab:subqa_quality_examples_zh_en} presents examples of some accepted and rejected sub-questions.
Specilfically, the LLM-Based verification based on the following criteria：
\begin{itemize}[leftmargin=15pt]
    \item \textbf{Answerability and Verifiability:} The sub-question must admit a concrete and finite set of plausible answers that can be independently verified, regardless of whether a single answer is enforced.
    \item \textbf{Specificity and Determinacy:} The sub-question should clearly target a specific fact or relation, avoiding vague references or ambiguous phrasing.
    \item \textbf{Temporal and Factual Stability:} The answer must reflect an objective, time-invariant fact that does not depend on evolving context or subjective interpretation.
\end{itemize}

\paragraph{Multi-Hop Question Composition}
After filtering invalid reasoning paths, we prompt the LLM to compose a coherent multi-hop question from each valid fact chain. 
Table~\ref{tab:multi_hop_composition_putin} provides an example of composing multi-hop question.
Specifically, the model replaces intermediate facts with appropriate referential expressions to ensure fluency and maintain a natural narrative flow across reasoning steps.  
The final question must contain only \textit{one explicit interrogative}, while earlier sub-questions should be embedded implicitly within the contextual description.  
The question should not reveal the answers to any intermediate steps, and all sub-parts must be logically connected through referential or causal relations to form a coherent reasoning chain.
To enhance the naturalness and readability of the composed question, appropriate \textit{contextual information} is added as needed.  
When any sub-question admits multiple valid reasoning trajectoies, we introduce additional constraints to disambiguate the reasoning path and ensure that the multi-hop question yields a \textbf{concrete and unique} final answer.

% We also retain the model’s evaluation outputs and rationales to support interpretability and guide potential human annotation or correction.

\begin{table*}[t]
\centering
\small
\begin{tabular}{p{0.45\linewidth} | p{0.5\linewidth}}
\toprule
\textbf{Seed Anchor: 普京 (Putin)} & \textbf{Composed Multi-hop Question} \\
\midrule
\textbf{$Q_1:$} 普京于哪一年首次当选俄罗斯总统？ \newline
\textit{In which year did Putin first become President of Russia?} \newline
\textbf{$A_1:$} 2000 \newline

\textbf{$Q_2:$} 哪一位科学家获得了2000年诺贝尔物理学奖？ \newline
\textit{Who won the Nobel Prize in Physics in 2000?} \newline
\textbf{$A_2:$} 阿尔费罗夫 / 克勒默 / 基尔比 (Alferov / Kroemer / Kilby) \newline

\textbf{$Q_3:$} 阿尔费罗夫逝世于哪一年？ \newline
\textit{In which year did Alferov pass away?} \newline
\textbf{$A_3:$} 2019
&
\textit{LLM composition:} \newline
普京首次当选俄罗斯总统的年份是\textcolor{red}{哪一年}，\textcolor{blue}{同年}哪一位科学家获得了诺贝尔物理学奖，\textcolor{red}{其}逝世年份是哪一年？ \newline
\textit{\textcolor{red}{When} did Putin first become President of Russia, \textcolor{red}{who} won the Nobel Prize in Physics \textcolor{blue}{in the same year}, and when did \textcolor{red}{he} pass away?} \newline

\textit{Human annotation:} \newline
普京首次当选俄罗斯总统的\textcolor{blue}{那一年}，\textcolor{blue}{一位科学家}获得了诺贝尔物理学奖。\textcolor{blue}{这位俄罗斯籍}科学家逝世于哪一年？ \newline
\textit{\textcolor{blue}{In the year} Putin was first elected President of Russia, \textcolor{blue}{a scientist} won the Nobel Prize in Physics, in which year did this \textcolor{blue}{Russian} scientist pass away?}
\\
\bottomrule
\end{tabular}
\caption{An example of composing a multi-hop question from a chain of sub-questions, based on the seed entity “普京 (Putin)”. \textbf{\textcolor{red}{Red} }indicates inappropriate wording, while \textbf{\textcolor{blue}{Blue}} denotes suitable revisions.}
\label{tab:multi_hop_composition_putin}
\vspace{-5pt}
\end{table*}

\subsection{Quality Control}  
We adopt a {human-in-the-loop} annotation process to ensure {factual accuracy} and {reasoning quality}, complementing LLM-based verification. Professional annotators are involved throughout the data construction pipeline, including seed validation, sub-question generation, and especially multi-hop question composition. Each instance is independently reviewed by two annotators, with disagreements resolved by a third. Annotators are provided with the \textit{complete data source}, including sub-questions, answers, LLM justifications, and final composed questions. They are instructed to verify all facts against \textit{authoritative sources} (e.g., Wikipedia, Baidu Baike). Each instance is evaluated based on the following criteria: \textbf{(1) Global answer uniqueness：} the reasoning chain must converge to a unique, concrete, and verifiable answer; \textbf{(2) Sequential consistency：} the steps must reflect genuine multi-hop inference without shortcuts or redundancy; \textbf{(3) Harmlessness：} all content must be free from harmful information or social bias.
Instances that do not meet these criteria are either revised or discarded. The final dataset has undergone LLM verification and human validation, supported by authoritative evidence sources.

\begin{table}[t]
  \centering
  \resizebox{0.95\linewidth}{!}{
    \begin{tabular}{lcc}
      \toprule
      \textbf{Statistics}  & \textbf{3-Hops} & \textbf{6-Hops} \\
      \midrule
      Initial Samples  & 1000 & 1000 \\
      \#  LLM-Generated Subquestions  & 1563 & 1164 \\
      \# Composed Multihop Sample  & 521 & 194 \\
      \# After Human Annotation & 480 & 166  \\
      Avg. \# Subquestion Length & 16.73 & 18.17  \\
      Avg. \# Subquestion Answer Length & 5.32 & 6.62  \\
      Avg. \# Whole Question Length & 39.19 & 68.51 \\
      Avg. \# Whole Answer Length & 4.85 & 6.48  \\
        Avg. \# Domain Coverage & 1.65 & 2.26  \\
      \bottomrule
    \end{tabular}
  }
  \caption{Overall statistics of \textsc{CCMOR}.}
  \label{tab:dataset_statistics}
  \vspace{-5pt}
\end{table}

\subsection{Dataset Statistics}
% Table~\ref{tab:dataset_statistics} details the construction and key properties of our multi-hop QA dataset. All samples are subject to multiple rounds of strict filtering, including LLM-based filtering and human annotation, ensuring the overall quality and reliability of the benchmark. Unlike typical commonsense datasets, each question in our collection frequently spans multiple domains, with 6-hop questions averaging 2.27 domains. This design facilitates robust evaluation of models' cross-domain reasoning abilities. Additionally, the dataset includes a substantial proportion of subquestions that require explicit reasoning and backtracking, making it a rigorous resource for assessing multi-step inference capabilities in large language models.

Table~\ref{tab:dataset_statistics} details the construction and key properties of our multi-hop QA dataset. All samples undergo multiple rounds of rigorous filtering, including LLM-based verification and human annotation, ensuring high quality and reliability. Unlike typical common-sense datasets, each question in our collection often spans multiple domains, with 6-hop questions averaging 2.26 domains, enabling a assessment of cross-domain reasoning capabilities.

Importantly, our dataset explicitly provides intermediate sub-questions and answers, allowing for fine-grained supervision of the reasoning process rather than merely evaluating final answers. This design not only enhances interpretability, but also enables targeted training and diagnosis of intermediate reasoning failures. Moreover, many sub-questions admit multiple plausible answers, requiring models to perform reasoning with backtracking to identify the correct path. These characteristics make our dataset a comprehensive benchmark for evaluating multi-step inference and traceable reasoning in large language models.

\begin{table*}[t]
\centering
\resizebox{0.96\textwidth}{!}{
\begin{tabular}{l|cc|cccc|cccc|c}
\toprule
\multirow{3}{*}{\textbf{Models}} &
\multicolumn{2}{c|}{\textbf{1-Hop}} &
\multicolumn{4}{c|}{\textbf{3-Hops}} &
\multicolumn{4}{c|}{\textbf{6-Hops}} &
\multirow{3}{*}{\textbf{Avg.}} \\
\cmidrule(lr){2-3} \cmidrule(lr){4-7} \cmidrule(lr){8-11}
& \multicolumn{2}{c|}{\textbf{OA}} 
& \multicolumn{2}{c}{\textbf{SQA}} & \multicolumn{2}{c|}{\textbf{OA}} 
& \multicolumn{2}{c}{\textbf{SQA}} & \multicolumn{2}{c|}{\textbf{OA}} \\
& \scriptsize{\textbf{Rouge-L}} & \scriptsize{\textbf{LLM-Judge}} 
& \scriptsize{\textbf{Rouge-L}} & \scriptsize{\textbf{LLM-Judge}} 
& \scriptsize{\textbf{Rouge-L}} & \scriptsize{\textbf{LLM-Judge}} 
& \scriptsize{\textbf{Rouge-L}} & \scriptsize{\textbf{LLM-Judge}} 
& \scriptsize{\textbf{Rouge-L}} & \scriptsize{\textbf{LLM-Judge}} \\
\midrule

\rowcolor{blue!5} 
\multicolumn{12}{c}{\textit{\textbf{System-1-Style Models}}} \\
\midrule
Qwen2.5-14B-Instruct & 53.91 & 41.22 & 62.69 & 55.95 & 44.93 & 29.02 & 63.65 & 57.84 & 39.32 & 25.83 & 32.02 \\
Qwen2.5-32B-Instruct & 56.68 & 43.98 & 66.18 & 62.63 & 50.74 & 41.22 & 67.32 & 60.38 & 41.18 & 30.46 & 38.55 \\
Qwen2.5-72B-Instruct & 64.19 & 54.44 & 72.76 & 67.06 & 60.23 & 48.96 & 75.14 & 68.40 & 55.00 & 43.71 & 49.04 \\
Qwen3-14B & 57.26 & 46.15 & 65.73 & 63.31 & 50.91 & 41.37 & 66.38 & 60.51 & 31.93 & 21.85 & 36.46 \\
Qwen3-32B & 61.69 & 46.75 & 67.20 & 63.39 & 52.17 & 43.15 & 69.97 & 63.87 & 39.58 & 27.15 & 39.02 \\
Qwen3-30A3B & 54.27 & 44.18 & 61.83 & 57.19 & 37.60 & 21.13 & 65.42 & 60.88 & 30.16 & 17.88 & 27.73 \\
Qwen3-235A22B & 66.28 & 61.74 & 74.12 & 72.46 & 60.09 & 51.19 & 76.43 & 74.56 & 48.15 & 41.33 & 51.42 \\
Yi-lightning & 69.63 & 67.85 & 76.97 & 76.10 & 65.90 & 61.01 & 80.50 & 79.42 & 62.44 & 54.30 & 61.05 \\
Moonshot-v1 & 62.47 & 55.34 & 76.39 & 72.45 & 52.69 & 36.76 & 78.08 & 74.03 & 45.79 & 31.79 & 41.30 \\
Baichuan4-Turbo & 63.09 & 68.05 & 73.03 & 80.49 & 55.56 & 43.30 & 72.62 & 80.55 & 37.01 & 26.49 & 45.95 \\
GLM-4-air & 69.57 & 68.18 & 81.30 & 82.72 & 55.73 & 46.58 & 84.40 & 85.29 & 53.01 & 45.95 & 53.57 \\
Doubao-1.5-pro & 65.17 & 65.48 & 82.04 & 78.87 & 69.25 & 63.84 & 83.42 & 80.13 & 60.01 & 53.64 & 60.99 \\
Deepseek-V3 & 65.15 & 69.82 & 82.39 & 82.28 & 72.01 & 72.77 & 84.41 & 83.92 & 71.61 & 64.24 & 68.94 \\
LlaMA3-70B-Instruct & 52.09 & 43.79 & 62.27 & 55.95 & 45.81 & 30.51 & 67.88 & 63.47 & 42.47 & 29.80 & 34.70 \\
GPT-4.1 & 68.50 & 62.33 & 75.23 & 72.36 & 61.12 & 48.21 & 78.98 & 76.57 & 62.14 & 54.30 & 54.95 \\
GPT-4o & 57.93 & 52.07 & 69.45 & 65.43 & 64.44 & 56.10 & 73.56 & 70.28 & 55.92 & 49.67 & 52.61 \\
Gemini-2.5-flash & 75.78 & 69.63 & 80.98 & 82.03 & 65.87 & 65.62 & 82.20 & 82.96 & 57.81 & 54.30 & 63.18 \\
\midrule
\rowcolor{blue!5} 
\multicolumn{12}{c}{\textit{\textbf{System-2-Style Models}}} \\
\midrule
Qwen3-14B-Think & 55.55 & 45.36 & 66.22 & 63.94 & 55.84 & 46.88 & 68.60 & 67.66 & 48.45 & 40.40 & 44.21 \\
Qwen3-32B-Think & 62.94 & 52.07 & 69.26 & 65.23 & 56.87 & 49.55 & 72.67 & 68.85 & 49.13 & 41.72 & 47.78 \\
Qwen3-30A3B-Think & 57.79 & 50.60 & 67.23 & 66.02 & 55.98 & 49.33 & 70.65 & 68.58 & 48.98 & 41.89 & 47.27 \\
Qwen3-235A22B-Think & 69.93 & 64.16 & 77.46 & 74.41 & 66.48 & 62.91 & 80.00 & 77.24 & 58.67 & 51.72 & 59.60 \\
Qwen-QwQ-32B & 61.38 & 53.65 & 67.94 & 65.46 & 55.87 & 46.43 & 71.91 & 69.98 & 49.76 & 40.40 & 46.83 \\
% openai-o3-mini & 56.92 &	52.07 & 63.36	& 62.83	&	55.82	&51.33	&	68.62&	 66.56	& 57.73 &	53.51& 52.30 \\
GLM-Z1-air & 71.86 & 65.81 & 83.84 & 82.55 & 58.79 & 49.11 & 85.15 & 85.51 & 46.21 & 37.33 & 50.75 \\
Doubao-1.5-Think & 72.50 & 67.65 & 81.20 & 81.71 & 71.47 & 68.45 & 82.15 & 82.01 & 67.81 & 62.25 & 66.12 \\
Deepseek-R1 & 76.66 & 75.15 & 85.98 & 85.91 & 78.91 & 75.89 & 86.80 & 86.87 & 71.72 & 66.89 & 72.64 \\
openai-o1 & 71.67 & 70.30 & 79.54 & 79.97 & 73.23 & 71.01 & 76.47 & 83.46 & 74.36 & 72.66 & 71.32 \\
Gemini-2.5-Pro & 77.55 & 77.91 & 82.96 & 85.12 & 74.17 & 73.38 & 85.87 & 87.47 & 71.98 & 69.54 & 73.61 \\
\midrule
\bottomrule
\end{tabular}
}
\caption{
Performance of baseline models on our proposed benchmark. \textbf{1-Hop} refers to results on single-hop seed questions. \textbf{3-Hops} and \textbf{6-Hops} correspond to multi-hop questions with increasing reasoning complexity. \textbf{SQA} and \textbf{OA} denote Stepwise Question Answering and Overall Answering settings, respectively. \textbf{Rouge-L} and \textbf{LLM-Judge} represent ROUGE-L recall and \textit{LLM-as-Judge} accuracy. \textbf{Avg.} reports the average \textit{LLM-as-Judge} accuracy for overall answering across all questions.
}
\label{tab:mainresults}
\end{table*}
\section{Experiments}

\subsection{Evaluated Models}
We evaluate a variety of mainstream large language models, categorized into two groups based on their reasoning paradigms: \textit{System-1-style} and \textit{System-2-style} models.

\textit{System-1-style} models rely on short chain-of-thought reasoning, favoring fast, intuitive, and heuristic-driven response. They typically produce concise answers with minimal intermediate reasoning steps. Representative models in this category include the Qwen2.5~\cite{qwen2.5}, Qwen3, and LLaMA~\cite{llama3modelcard} series, among other widely used LLMs~\cite{deepseekai2025deepseekv3technicalreport,wake2025yilightningtechnicalreport,lin2024baichuanalignmenttechnicalreport,gemini25propreview}.

% such as DeepSeek-v3~\cite{liu2024deepseek}, Doubao-1.5-Pro~\cite{doubao15pro2025} and GPT-4o~\cite{gpt-4o}, 
In contrast, recently emerging \textit{System-2-style} models such as DeepSeek-R1~\cite{guo2025deepseek}, OpenAI-o1~\cite{jaech2024openai}, and Qwen-QwQ~\cite{team2025qwq} adopt long chain-of-thought reasoning, characterized by more deliberate and structured analytical processes. This reasoning paradigm aligns with the principles of \textit{System 2} thinking~\cite{evans2003two, kannengiesser2019design}, which emphasizes slow, reflective, and logically grounded cognition. These models are typically trained to explicitly generate intermediate reasoning steps, thereby improving both answer accuracy and interpretability.
% While LRMs generally achieve higher accuracy on complex reasoning tasks, they tend to produce longer and slower responses as a trade-off.

\subsection{Evaluation}

We evaluate model performance from two complementary perspectives:

\paragraph{\textbf{Stepwise Question Answering (SQA):}}  
In this setting, each multi-hop question is decomposed into sub-questions.
The model is prompted to answer them respectively, with the reference answer from the previous step substituted into subsequent sub-questions to eliminate coreference ambiguity.
This setting assesses the model’s ability to address each component of a multi-hop reasoning chain, reflecting its factual recall capabilities.

\paragraph{\textbf{Overall Answering (OA):}} In this setting, the model is presented with the complete multi-hop question and tasked with producing the final answer. This setup evaluates not only the model’s ability to recall knowledge for each implicit sub-question, but also its capacity to integrate these steps into a coherent reasoning process coherently.

For both SQA and OA, we adopt two evaluation metrics: \textbf{Rouge-L Recall} and \textbf{\textit{LLM-as-Judge} Accuracy}.  
Rouge-L recall measures the lexical overlap between model output and gold answer, serving as an automatic indicator of surface-level correctness. Since we observe that models tend to generate redundant content in their answers, we use recall rather than f1 score.
\textit{LLM-as-Judge} accuracy offers a semantic-level evaluation by leveraging three independent judge models to assess the alignment of predicted answers with the reference in terms of meaning and reasoning validity. The final decision is determined by majority voting among the three judges. Detailed evaluation settings and prompt templates are provided in the Appendix~\ref{sec:prompt}.

% In both evaluation settings, we apply chain-of-thought prompting to guide models to produce outputs that conform to predefined templates. This structured output design facilitates reliable automatic extraction and evaluation. Detailed prompt templates and formatting instructions are provided in the Appendix.

\subsection{Main Results}
Table~\ref{tab:mainresults} summarizes the performance of baseline models on our benchmark. Models such as \texttt{DeepSeek-R1}, \texttt{OpenAI-o1}, \texttt{Doubao-1.5-Pro}, and \texttt{Gemini-2.5-Pro} exhibit strong performance, particularly on more complex multi-hop questions.
However, even top-performing models score below 75\% in average multi-hop accuracy, highlighting the overall difficulty of our benchmark.

Closed-source models generally outperform open-source ones, likely due to larger model sizes and more extensive training data. Within the same model family (e.g., Qwen2.5 and Qwen3), larger variants consistently achieve better results.
Chinese community models such as \texttt{Yi-lightning}, \texttt{GLM-4}, and \texttt{Baichuan4-Turbo} perform particularly well in the SQA setting, suggesting that language adaptation and corpus coverage are critical for Chinese commonsense QA task.
System-2-style models (e.g., \texttt{DeepSeek-R1}, \texttt{Doubao-1.5-Think}, and \texttt{Gemini-2.5-Pro}) consistently outperform their System-1-style counterparts (e.g., \texttt{DeepSeek-V3}, \texttt{Doubao-1.5-Pro}, and \texttt{Gemini-2.5-Flash}) in the OA setting, indicating that structured reasoning is more effective for multi-hop tasks. Moreover, performance declines significantly as the number of reasoning hops increases. The persistent performance gap between SQA and OA across all models suggests that, although models can handle sub-questions well, synthesizing intermediate answers into a final correct response remains challenging.

\begin{figure}[t]
    \centering
    \includegraphics[width=1\linewidth]{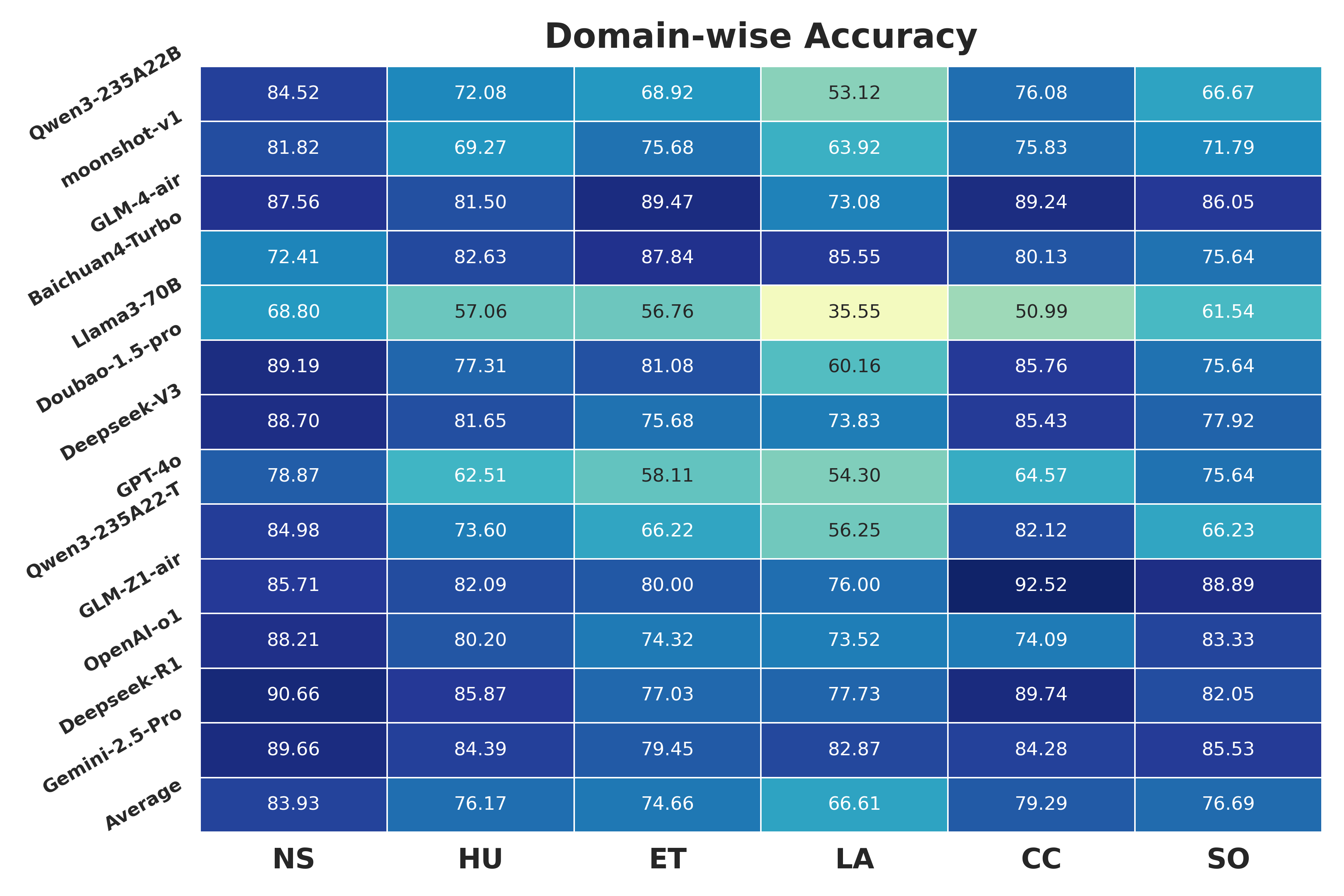}
    \caption{
        Domain-wise \textit{LLM-as-Judge} accuracy for different models. \textbf{CC, HU, ET, LA, SO} and \textbf{NS} represent “Chinese Culture”, “Humanities”, “Engineering and Technology”, “Life and Art”, “Society”, and “Natural Science”, respectively.
    }
    \label{fig:domain-heatmap}
\end{figure}

\begin{figure}[t]
    \centering
    \includegraphics[width=1\linewidth]{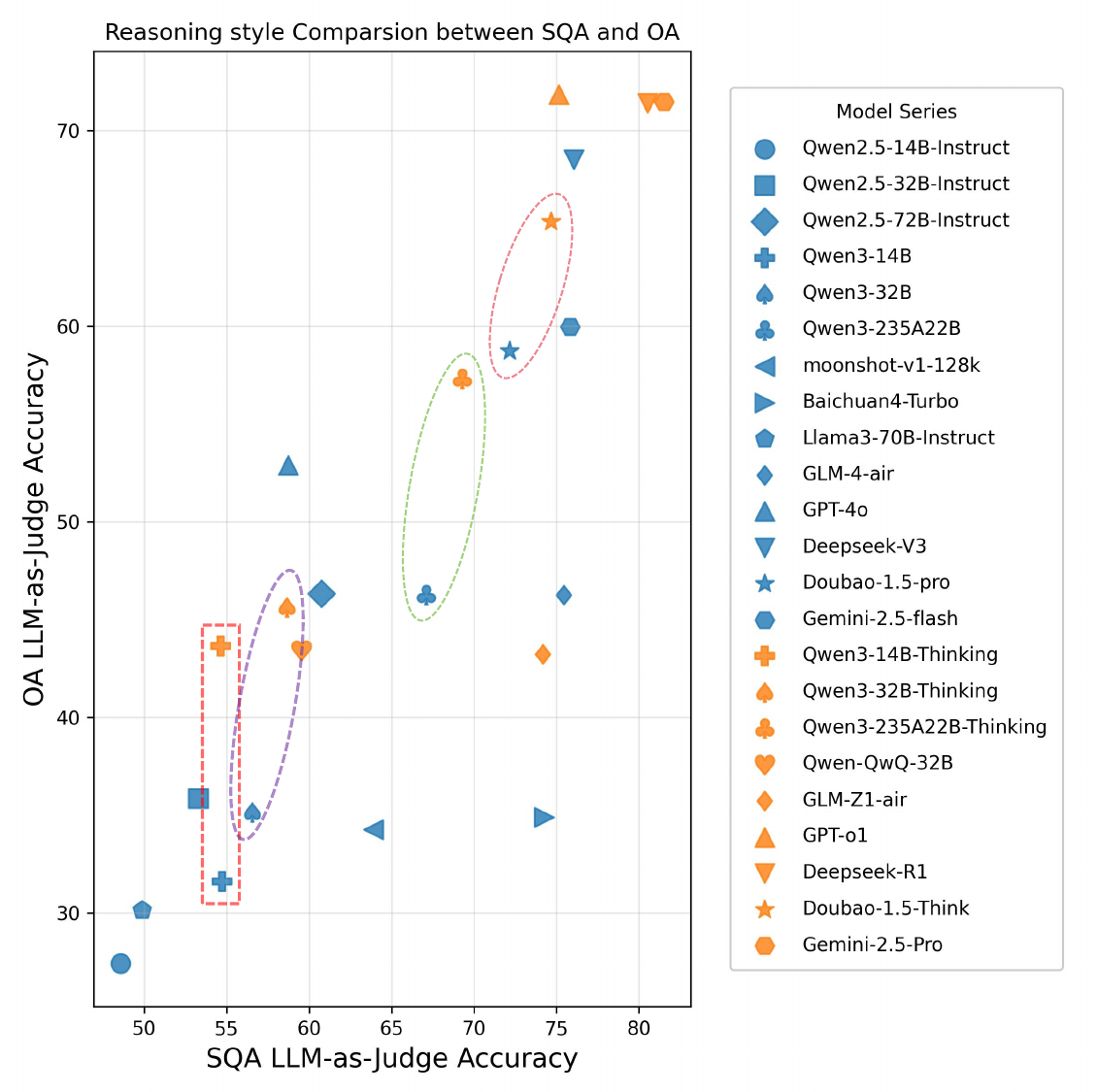}
    \caption{
Performance of models with different reasoning styles in the sub-question answering (SQA) and overall answering (OA) settings. \textcolor{blue}{Blue} represent system-1-style models, while \textcolor{orange}{Orange} represent system-2-style models.}
\label{fig:reasoning-style}
\end{figure}

\begin{figure}[t]
    \centering
    \includegraphics[width=0.96\linewidth]{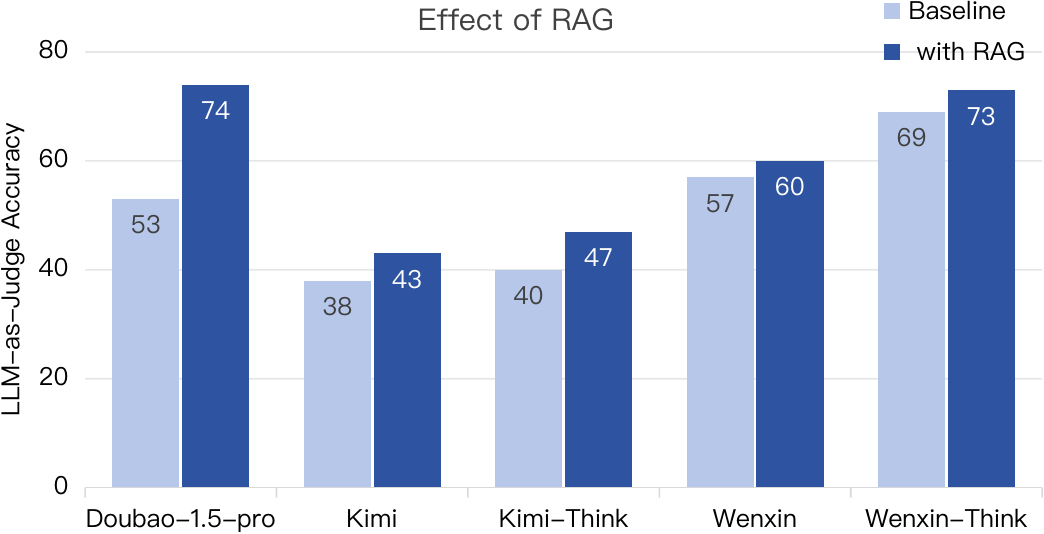}
    \caption{
        \textit{LLM-as-Judge} accuracy of different baselinse models with RAG.
    }
    \label{fig:rag_bar}
\end{figure}

\section{Analysis}    
\subsection{Domain-wise Performance}  
Figure~\ref{fig:domain-heatmap} illustrates the performance of representative models across different domains. Most models achieve strong results in the \textit{Natural Science} domain, with an average score of 83.93, while performance in the \textit{Life and Art} domain is notably lower, averaging 66.61. This suggests that fact-centric domains are generally more manageable for current models, whereas domains requiring procedural or abstract reasoning remain more challenging.

Moreover, Chinese community models such as \texttt{GLM-4}, \texttt{Doubao-1.5-Pro}, and the \texttt{DeepSeek} series tend to outperform others in the \textit{Chinese Culture} domain, likely due to domain-specific training on Chinese data. Overall, the relatively small performance gap across domains suggests that our dataset is well-balanced in difficulty and does not exhibit significant domain-specific bias.

\subsection{Reasoning-style Comparsion}
We explore the performance of different reasoning styles on both single-hop and multi-hop tasks, and investigate whether long-CoT reasoning provides measurable benefits for complex multi-hop questions. As shown in Figure~\ref{fig:reasoning-style}, models employing system-2-style reasoning consistently outperform system-1-style models in both the sub-question answering and overall answering settings. This indicates that deliberate reasoning can be beneficial even for commonsense tasks that are primarily involve factual recall.

A more notable contrast emerges when comparing different reasoning modes of the same model, such as Qwen-3 series, which is highlighted with dashed boxes in the Figure~\ref{fig:reasoning-style}.
Although both modes achieve comparable performance in the SQA setting, suggesting similar capabilities in factual recall, the \textit{think mode} demonstrates a significant improvement in OA scores. This suggests that explicit reasoning steps enhance the model’s ability to integrate the memory of individual sub-questions into a coherent multi-step reasoning chain, ultimately leading to more accurate final answers.

% \paragraph{Chinese vs. English-centric Models}
% Chinese-native models generally outperform English-centric models in culturally specific domains such as \textit{Chinese Culture} and \textit{Society}, reflecting better linguistic alignment and domain adaptation.

% \paragraph{Open-source vs. Close-source Models}
% Proprietary models like GPT-4 and Claude achieve higher overall accuracy, especially on complex reasoning chains. However, leading open-source models demonstrate competitive results in memorization-heavy domains, showing potential for domain-specialized tasks.

% \paragraph{System 1 vs. System 2 Models}
% We contrast direct-answering models with those capable of multi-step, deliberative reasoning (e.g., via CoT or tool use). The latter type demonstrates clear advantages in deep multi-hop questions and temporal or multi-entity reasoning tasks.

% \paragraph{Intra-domain vs. Cross-domain Reasoning}
% Accuracy drops significantly when questions involve reasoning across domains rather than within a single domain. This highlights a key limitation of current models in transferring knowledge and maintaining coherence in multi-hop reasoning across heterogeneous topics.

\subsection{Prompting Strategies}
We compare the impact of different prompting strategies. Experiments are conducted on a subset consisting of 200 3-hops and 100 6-hops questions. We evaluate a set of Chinese community models ( Deepseek-v3 and Doubao-1.5-pro ) and a set of English community models ( Gemini-2.5-flash- and GPT-4o ), using the average \textit{LLM-as-Judge} accuracy under different prompting settings as the evaluation metric.
Specifically, we compare: (1) \textit{\textbf{Direct}} and \textit{\textbf{CoT}} prompting, where the CoT prompt guides the model perform step-by-step reasoning; (2) \textit{\textbf{Chinese}} and \textit{\textbf{English}} prompting, where the origin chinese prompt are translated into english;  (3) \textit{\textbf{Zero-shot}} and  \textit{\textbf{Few-shot}} prompting, where the few-shot prompting includes an example to illustrate multi-hop reasoning.

As shown in Table~\ref{tab:prompt-strategies}, prompt language (Chinese vs. English) and the use of few-shot examples have limited impact on accuracy. In contrast, introducing heuristic, guided reasoning via chain-of-thought (CoT) prompts yields more substantial improvements. This indicates that commonsense multi-hop reasoning tasks are relatively robust to superficial prompt settings but benefit from prompts that explicitly guide the reasoning process.

\begin{table}[t]
\centering
\small
\resizebox{\linewidth}{!}{
\begin{tabular}{l|cc}
\toprule
\textbf{Prompt Strategy} & \textbf{CN-LLMs} & \textbf{EN-LLMs} \\
\midrule
\texttt{Direct | Chinese | Zero-shot} & 62.5 & 57.5 \\
\texttt{Direct | Chinese | Few-shot}  & 62.0 & 59.5 \\
\texttt{Direct | English | Zero-shot} & 64.5 & 60.0 \\
\texttt{Direct | English | Few-shot}  & 61.5 & 56.0 \\
\hspace{0.6em} \texttt{CoT \hspace{0.2em}   | Chinese | Zero-shot} & 65.5 & 60.5 \\
\hspace{0.6em} \texttt{CoT \hspace{0.2em}     | Chinese | Few-shot}  & 65.5 & 61.0 \\
\hspace{0.6em} \texttt{CoT \hspace{0.2em}    | English | Zero-shot} & 61.0 & 61.5 \\
\hspace{0.6em} \texttt{CoT \hspace{0.2em}     | English | Few-shot}  & 63.5 & 60.5 \\
\bottomrule
\end{tabular}
}
\caption{\textit{LLM-as-Judge} accuracy with different prompt strategies.}
\label{tab:prompt-strategies}
\vspace{-5pt}
\end{table}

\subsection{Effect of Retrieval-Augmented Generation}
We investigate the impact of retrieval-augmented generation (RAG) on enhancing multi-hop reasoning capabilities. Experiments are conducted on a subset of 50 three-hop and 50 six-hop questions using five models: \texttt{Doubao-1.5-Pro}, \texttt{Kimi}, \texttt{Kimi-Think}, \texttt{Wenxin}, and \texttt{Wenxin-Think}. For each model, retrieval is implemented via its official interface. “Think” variants denote configurations with the “deep thinking” option enabled, while all other settings remain at their default values.

As shown in Figure~\ref{fig:rag_bar}, integrating RAG consistently yields substantial improvements across all models, with an average accuracy gain of approximately 9.5 percentage points. However, the magnitude of improvement varies significantly across models. For instance, Kimi and Wenxin show relatively limited gains. Our analysis reveals that in these cases, the retrieved content often lacks relevant information, leading the models to reject answer even for questions they could answer correctly without retrieval. In contrast, Doubao demonstrates significant improvements, likely due to its adaptive utilization of retrieved content and support for multi-turn retrieval, which is particularly advantageous for multi-hop reasoning.
We provide detailed examples in Appendix~\ref{sec:appendix_case_model}.
% ~\ref{appendix:rag_case_study}.

\subsection{Agreement between \textit{LLM-as-Judge} and Human Evaluation}
Table~\ref{table:human_and_llm} reports the Cohen's Kappa scores measuring agreement between human annotators and the \textit{LLM-as-Judge} evaluation for both single-hop and multi-hop tasks, using Deepseek-v3, Doubao-1.5-pro and GPT-4o as evaluators. In all cases, the Cohen's Kappa score exceeds 95\%, indicating almost perfect agreement between LLM-based and human judgments. By adopting majority voting from three independent LLM judges, we mitigate model-specific bias and strengthen the robustness and credibility of \textit{LLM-as-Judge} evaluation.
\begin{table}[t]
    \centering
    \resizebox{\linewidth}{!}{
    \begin{tabular}{lccc}
        \hline
        Level & \multicolumn{3}{c}{Agreement ($\kappa\uparrow$)} \\
        \cline{2-4}
         & Doubao-1.5-pro & Deepseek-v3 & GPT-4o  \\
        \hline
        Single-hop & 95.6 & 96.4 & 95.6  \\
        Multi-hops & 97.7 & 95.3 & 96.8  \\
        \hline
    \end{tabular}
    }
        \caption{Cohen's Kappa agreement ($\kappa$) between human evaluation and \textit{LLM-as-Judge}.}
        \label{table:human_and_llm}
\vspace{0em}
\end{table}

\section{Related Works}
\subsection{Multi-hop Reasoning Benchmarks}
The development of multi-hop reasoning benchmarks evolves from early Wikipedia-based datasets to more advanced and specialized evaluations. The seminal \textit{HotpotQA}~\cite{yang2018hotpotqa} introduces core paradigms for evidence-based reasoning, while \textit{2WikiMultiHopQA}~\cite{xanh2020_2wikimultihop} applies adversarial filtering to improve robustness. Subsequent benchmarks target specific challenges: \textit{MuSiQue}~\cite{trivedi2022interleavingretrieval} promotes verifiable reasoning through question decomposition, \textit{Multihop-RAG}~\cite{tang2024multihop} evaluates the integration of retrieval in multi-hop generation, and \textit{MQuAKE}~\cite{zhong-etal-2023-mquake} focuses on knowledge editing via multi-hop questions.
Recent works~\cite{schnitzler2024morehopqa,wu2024cofca,zhu2024fanoutqa,veuthey2025meqametaevaluationframeworkquestion,zhou2025reasoning} leverages LLMs to construct higher-quality multi-hop questions and extend benchmark coverage to multimodal reasoning tasks.
However, Chinese multi-hop reasoning remains underexplored. While efforts such as \textit{NLPCC-MH}~\cite{NLPCC-MH}, CoreCode~\cite{shi2024corecode}, and CHARM~\cite{sun2024benchmarking} represent initial progress by incorporating cultural knowledge into Chinese multi-hop datasets, they do not systematically support verifiable multi-step reasoning through explicit decomposition. These limitations highlight the need for native Chinese benchmarks that combine cultural relevance with explicitly decomposed, verifiable reasoning steps, and adopt robust design methodologies to minimize annotation bias.

\subsection{Chinese Commonsense Benchmarks}
The development of Chinese commonsense reasoning benchmarks has evolved through multiple stages, yet significant gaps persist compared to their English counterparts. Early efforts primarily involved translating existing English benchmarks~\cite{conneau2018xnli,ponti2020xkglue}. Subsequent work introduced native Chinese evaluations with varying degrees of commonsense coverage, including general language understanding tests~\cite{xu2020clue,li2023cmmlu}, logical reasoning assessments~\cite{liu2020logiqa,zhong2023legal}.
Benchmarks such as \textit{C3}~\cite{sun2019investigatingpriorknowledgechallenging} and \textit{CMQA}~\cite{ju-etal-2022-cmqa} , have further advanced the field by introducing more complex and diverse question formats. Among existing benchmarks, \textit{Chinese SimpleQA}~\cite{he2024chinesesimpleqachinesefactuality} stands out for its comprehensive coverage and high-quality construction. Each question undergoes a rigorous validation process involving multiple large language models and human annotators to ensure quality and cultural appropriateness.
However, existing Chinese SimpleQA benchmarks primarily focus on single-hop, fact-based questions, which fall short in evaluating the advanced reasoning capabilities of large language models. Our proposed \textsc{CCMOR} extends existing Chinese QA resources into multi-hop questions. It enables more effective benchmarking of state-of-the-art models on Chinese multi-hop scenario, while providing verifiable intermediate reasoning steps for process-level evaluation.

\section{Conclusion}
We present a novel benchmark for evaluating Chinese multi-hop commonsense reasoning. The dataset is constructed via an LLM-driven generation pipeline, followed by expert human verification to ensure both quality and coverage. It explicitly targets the integration of Chinese cultural knowledge, verifiable intermediate reasoning steps, and factual information across diverse domains.

Our empirical analysis reveals that, despite strong sub-question answering performance, state-of-the-art models still struggle with knowledge-intensive and compositional reasoning, especially in long-tail scenarios. We further analyze the impact of different reasoning styles, prompting strategies, and the incorporation of retrieval augmentation, demonstrating their varying effectiveness in enhancing multi-hop performance.
This benchmark fills a critical gap in the landscape of Chinese multi-hop commonsense reasoning datasets, offering a rigorous and culturally grounded resource for evaluating and advancing capabilities in LLMs.

\section*{Limitations}
While our benchmark provides a valuable resource for evaluating Chinese multi-hop commonsense reasoning, it has several limitations. 
\begin{itemize}[leftmargin=15pt]
    \item The dataset construction process depends on LLM-generated question-answer pairs, which may introduce hallucinations or inherent biases from the models. Although we adopt round-robin strategy to query multiple LLMs and employ expert human verification, some subtle errors or inconsistencies in reasoning may persist.

    \item The evaluation of model performance relies on \textit{LLM-as-a-judge}, which, while effective for scalable open-ended assessment, may be influenced by model-specific preferences or limitations in understanding complex reasoning. Although we supplement this with Rouge-L Recall for additional reference, such lexical metrics may not fully reflect reasoning accuracy or semantic correctness.

    \item Our benchmark focuses on multi-hop reasoning based on textual commonsense knowledge. Scenarios requiring multimodal reasoning, procedural understanding, or interactive dialogue are not covered in the current dataset and are left for future exploration.
\end{itemize}
We hope \textsc{CCMOR} will support future research in multilingual reasoning, dataset development, and the design of reasoning-specialized models.

\section*{Ethical Considerations}
We employ human annotators to construct and verify our dataset. They are recruited through a professional annotation platform and are fairly compensated in accordance with local standards. Prior to annotation, all annotators review detailed task guidelines and prompt templates in Appendix~\ref{sec:prompt}, where data usage policies and content safety requirements are clearly specified.

All annotators provide informed consent and are explicitly instructed to avoid generating content that involves bias, private information, or ethical risks. Annotators are professional language workers based in China, with relevant experience and no conflicts of interest.

\bibliography{main}

\begin{thebibliography}{36}
\providecommand{\natexlab}[1]{#1}

\bibitem[{AI@Meta(2024)}]{llama3modelcard}
AI@Meta. 2024.
\newblock \href {https://github.com/meta-llama/llama3/blob/main/MODEL_CARD.md} {Llama 3 model card}.

\bibitem[{Conneau et~al.(2018)Conneau, Rinott, Lample, Williams, Bowman, Schwenk, and Stoyanov}]{conneau2018xnli}
Alexis Conneau, Ruty Rinott, Guillaume Lample, Adina Williams, Samuel~R Bowman, Holger Schwenk, and Veselin Stoyanov. 2018.
\newblock Xnli: Evaluating cross-lingual sentence representations.
\newblock \emph{arXiv preprint arXiv:1809.05053}.

\bibitem[{DeepSeek-AI et~al.(2025)DeepSeek-AI, Liu, Feng, Xue, Wang, Wu, Lu, Zhao, Deng, Zhang, Ruan, Dai, Guo, Yang, Chen, Ji, Li, Lin, Dai, Luo, Hao, Chen, Li, Zhang, Bao, Xu, Wang, Zhang, Ding, Xin, Gao, Li, Qu, Cai, Liang, Guo, Ni, Li, Wang, Chen, Chen, Yuan, Qiu, Li, Song, Dong, Hu, Gao, Guan, Huang, Yu, Wang, Zhang, Xu, Xia, Zhao, Wang, Zhang, Li, Wang, Zhang, Zhang, Tang, Li, Tian, Huang, Wang, Zhang, Wang, Zhu, Chen, Du, Chen, Jin, Ge, Zhang, Pan, Wang, Xu, Zhang, Chen, Li, Lu, Zhou, Chen, Wu, Ye, Ye, Ma, Wang, Zhou, Yu, Zhou, Pan, Wang, Yun, Pei, Sun, Xiao, Zeng, Zhao, An, Liu, Liang, Gao, Yu, Zhang, Li, Jin, Wang, Bi, Liu, Wang, Shen, Chen, Zhang, Chen, Nie, Sun, Wang, Cheng, Liu, Xie, Liu, Yu, Song, Shan, Zhou, Yang, Li, Su, Lin, Li, Wang, Wei, Zhu, Zhang, Xu, Xu, Huang, Li, Zhao, Sun, Li, Wang, Yu, Zheng, Zhang, Shi, Xiong, He, Tang, Piao, Wang, Tan, Ma, Liu, Guo, Wu, Ou, Zhu, Wang, Gong, Zou, He, Zha, Xiong, Ma, Yan, Luo, You, Liu, Zhou, Wu, Ren, Ren, Sha, Fu, Xu, Huang, Zhang, Xie, Zhang, Hao,
  Gou, Ma, Yan, Shao, Xu, Wu, Zhang, Li, Gu, Zhu, Liu, Li, Xie, Song, Gao, and Pan}]{deepseekai2025deepseekv3technicalreport}
DeepSeek-AI, Aixin Liu, Bei Feng, Bing Xue, Bingxuan Wang, Bochao Wu, Chengda Lu, Chenggang Zhao, Chengqi Deng, Chenyu Zhang, Chong Ruan, Damai Dai, Daya Guo, Dejian Yang, Deli Chen, Dongjie Ji, Erhang Li, Fangyun Lin, Fucong Dai, Fuli Luo, Guangbo Hao, Guanting Chen, Guowei Li, H.~Zhang, Han Bao, Hanwei Xu, Haocheng Wang, Haowei Zhang, Honghui Ding, Huajian Xin, Huazuo Gao, Hui Li, Hui Qu, J.~L. Cai, Jian Liang, Jianzhong Guo, Jiaqi Ni, Jiashi Li, Jiawei Wang, Jin Chen, Jingchang Chen, Jingyang Yuan, Junjie Qiu, Junlong Li, Junxiao Song, Kai Dong, Kai Hu, Kaige Gao, Kang Guan, Kexin Huang, Kuai Yu, Lean Wang, Lecong Zhang, Lei Xu, Leyi Xia, Liang Zhao, Litong Wang, Liyue Zhang, Meng Li, Miaojun Wang, Mingchuan Zhang, Minghua Zhang, Minghui Tang, Mingming Li, Ning Tian, Panpan Huang, Peiyi Wang, Peng Zhang, Qiancheng Wang, Qihao Zhu, Qinyu Chen, Qiushi Du, R.~J. Chen, R.~L. Jin, Ruiqi Ge, Ruisong Zhang, Ruizhe Pan, Runji Wang, Runxin Xu, Ruoyu Zhang, Ruyi Chen, S.~S. Li, Shanghao Lu, Shangyan Zhou, Shanhuang
  Chen, Shaoqing Wu, Shengfeng Ye, Shengfeng Ye, Shirong Ma, Shiyu Wang, Shuang Zhou, Shuiping Yu, Shunfeng Zhou, Shuting Pan, T.~Wang, Tao Yun, Tian Pei, Tianyu Sun, W.~L. Xiao, Wangding Zeng, Wanjia Zhao, Wei An, Wen Liu, Wenfeng Liang, Wenjun Gao, Wenqin Yu, Wentao Zhang, X.~Q. Li, Xiangyue Jin, Xianzu Wang, Xiao Bi, Xiaodong Liu, Xiaohan Wang, Xiaojin Shen, Xiaokang Chen, Xiaokang Zhang, Xiaosha Chen, Xiaotao Nie, Xiaowen Sun, Xiaoxiang Wang, Xin Cheng, Xin Liu, Xin Xie, Xingchao Liu, Xingkai Yu, Xinnan Song, Xinxia Shan, Xinyi Zhou, Xinyu Yang, Xinyuan Li, Xuecheng Su, Xuheng Lin, Y.~K. Li, Y.~Q. Wang, Y.~X. Wei, Y.~X. Zhu, Yang Zhang, Yanhong Xu, Yanhong Xu, Yanping Huang, Yao Li, Yao Zhao, Yaofeng Sun, Yaohui Li, Yaohui Wang, Yi~Yu, Yi~Zheng, Yichao Zhang, Yifan Shi, Yiliang Xiong, Ying He, Ying Tang, Yishi Piao, Yisong Wang, Yixuan Tan, Yiyang Ma, Yiyuan Liu, Yongqiang Guo, Yu~Wu, Yuan Ou, Yuchen Zhu, Yuduan Wang, Yue Gong, Yuheng Zou, Yujia He, Yukun Zha, Yunfan Xiong, Yunxian Ma, Yuting Yan, Yuxiang
  Luo, Yuxiang You, Yuxuan Liu, Yuyang Zhou, Z.~F. Wu, Z.~Z. Ren, Zehui Ren, Zhangli Sha, Zhe Fu, Zhean Xu, Zhen Huang, Zhen Zhang, Zhenda Xie, Zhengyan Zhang, Zhewen Hao, Zhibin Gou, Zhicheng Ma, Zhigang Yan, Zhihong Shao, Zhipeng Xu, Zhiyu Wu, Zhongyu Zhang, Zhuoshu Li, Zihui Gu, Zijia Zhu, Zijun Liu, Zilin Li, Ziwei Xie, Ziyang Song, Ziyi Gao, and Zizheng Pan. 2025.
\newblock \href {https://arxiv.org/abs/2412.19437} {Deepseek-v3 technical report}.
\newblock \emph{Preprint}, arXiv:2412.19437.

\bibitem[{Dua et~al.(2019)Dua, Wang, Dasigi, Stanovsky, Singh, and Gardner}]{dua2019drop}
Dheeru Dua, Yizhong Wang, Pradeep Dasigi, Gabriel Stanovsky, Sameer Singh, and Matt Gardner. 2019.
\newblock Drop: A reading comprehension benchmark requiring discrete reasoning over paragraphs.
\newblock \emph{arXiv preprint arXiv:1903.00161}.

\bibitem[{Evans(2003)}]{evans2003two}
Jonathan St~BT Evans. 2003.
\newblock In two minds: dual-process accounts of reasoning.
\newblock \emph{Trends in cognitive sciences}, 7(10):454--459.

\bibitem[{Google(2025)}]{gemini25propreview}
Google. 2025.
\newblock \href {https://storage.googleapis.com/model-cards/documents/gemini-2.5-pro-preview.pdf} {Gemini 2.5 pro preview model card}.

\bibitem[{Guo et~al.(2025)Guo, Yang, Zhang, Song, Zhang, Xu, Zhu, Ma, Wang, Bi et~al.}]{guo2025deepseek}
Daya Guo, Dejian Yang, Haowei Zhang, Junxiao Song, Ruoyu Zhang, Runxin Xu, Qihao Zhu, Shirong Ma, Peiyi Wang, Xiao Bi, et~al. 2025.
\newblock Deepseek-r1: Incentivizing reasoning capability in llms via reinforcement learning.
\newblock \emph{arXiv preprint arXiv:2501.12948}.

\bibitem[{He et~al.(2024)He, Li, Liu, Tan, Wang, Huang, Bu, Guo, Hu, Zheng, Lin, Liu, Sun, Lin, Zheng, Zhu, Su, and Zheng}]{he2024chinesesimpleqachinesefactuality}
Yancheng He, Shilong Li, Jiaheng Liu, Yingshui Tan, Weixun Wang, Hui Huang, Xingyuan Bu, Hangyu Guo, Chengwei Hu, Boren Zheng, Zhuoran Lin, Xuepeng Liu, Dekai Sun, Shirong Lin, Zhicheng Zheng, Xiaoyong Zhu, Wenbo Su, and Bo~Zheng. 2024.
\newblock \href {https://arxiv.org/abs/2411.07140} {Chinese simpleqa: A chinese factuality evaluation for large language models}.
\newblock \emph{Preprint}, arXiv:2411.07140.

\bibitem[{Ho et~al.(2020)Ho, Duong~Nguyen, Sugawara, and Aizawa}]{xanh2020_2wikimultihop}
Xanh Ho, Anh-Khoa Duong~Nguyen, Saku Sugawara, and Akiko Aizawa. 2020.
\newblock \href {https://www.aclweb.org/anthology/2020.coling-main.580} {Constructing a multi-hop {QA} dataset for comprehensive evaluation of reasoning steps}.
\newblock In \emph{Proceedings of the 28th International Conference on Computational Linguistics}, pages 6609--6625, Barcelona, Spain (Online). International Committee on Computational Linguistics.

\bibitem[{Jaech et~al.(2024)Jaech, Kalai, Lerer, Richardson, El-Kishky, Low, Helyar, Madry, Beutel, Carney et~al.}]{jaech2024openai}
Aaron Jaech, Adam Kalai, Adam Lerer, Adam Richardson, Ahmed El-Kishky, Aiden Low, Alec Helyar, Aleksander Madry, Alex Beutel, Alex Carney, et~al. 2024.
\newblock Openai o1 system card.
\newblock \emph{arXiv preprint arXiv:2412.16720}.

\bibitem[{Ju et~al.(2022)Ju, Wang, Zhang, Zheng, Liu, and Zhao}]{ju-etal-2022-cmqa}
Yiming Ju, Weikang Wang, Yuanzhe Zhang, Suncong Zheng, Kang Liu, and Jun Zhao. 2022.
\newblock \href {https://aclanthology.org/2022.coling-1.146/} {{CMQA}: A dataset of conditional question answering with multiple-span answers}.
\newblock In \emph{Proceedings of the 29th International Conference on Computational Linguistics}, pages 1697--1707, Gyeongju, Republic of Korea. International Committee on Computational Linguistics.

\bibitem[{Kannengiesser and Gero(2019)}]{kannengiesser2019design}
Udo Kannengiesser and John~S Gero. 2019.
\newblock Design thinking, fast and slow: A framework for kahneman’s dual-system theory in design.
\newblock \emph{Design Science}, 5:e10.

\bibitem[{Li et~al.(2023)Li, Zhang, Koto, Yang, Zhao, Gong, Duan, and Baldwin}]{li2023cmmlu}
Haonan Li, Yixuan Zhang, Fajri Koto, Yifei Yang, Hai Zhao, Yeyun Gong, Nan Duan, and Timothy Baldwin. 2023.
\newblock \href {https://arxiv.org/abs/2306.09212} {Cmmlu: Measuring massive multitask language understanding in chinese}.
\newblock \emph{Preprint}, arXiv:2306.09212.

\bibitem[{Lin et~al.(2024)Lin, Yang, Shen, Sun, Li, Zhang, Zhu, Zhang, Zheng, Li, Zhou, Chen, Qin, Li, Liang, Li, Li, Wang, Dong, Fang, Xu, Cui, Zhang, Zhou, and Chen}]{lin2024baichuanalignmenttechnicalreport}
Mingan Lin, Fan Yang, Yanjun Shen, Haoze Sun, Tianpeng Li, Tao Zhang, Chenzheng Zhu, Tao Zhang, Miao Zheng, Xu~Li, Yijie Zhou, Mingyang Chen, Yanzhao Qin, Youquan Li, Hao Liang, Fei Li, Yadong Li, Mang Wang, Guosheng Dong, Kun Fang, Jianhua Xu, Bin Cui, Wentao Zhang, Zenan Zhou, and Weipeng Chen. 2024.
\newblock \href {https://arxiv.org/abs/2410.14940} {Baichuan alignment technical report}.
\newblock \emph{Preprint}, arXiv:2410.14940.

\bibitem[{Liu et~al.(2020)Liu, Cui, Liu, Wang, Hu, and Zhang}]{liu2020logiqa}
Zheng Liu, Yiming Cui, Wanxiang Liu, Shijin Wang, Guoping Hu, and Yue Zhang. 2020.
\newblock Logiqa: A challenge dataset for machine reading comprehension with logical reasoning.
\newblock \emph{arXiv preprint arXiv:2007.08124}.

\bibitem[{Ponti et~al.(2020)Ponti, Vuli{\'c}, Glava{\v{s}}, Reichart, and Korhonen}]{ponti2020xkglue}
Edoardo~Maria Ponti, Ivan Vuli{\'c}, Goran Glava{\v{s}}, Roi Reichart, and Anna Korhonen. 2020.
\newblock Xcopa: A multilingual dataset for causal commonsense reasoning.
\newblock In \emph{Proceedings of the 2020 Conference on Empirical Methods in Natural Language Processing (EMNLP)}, pages 2362--2376.

\bibitem[{Schnitzler et~al.(2024)Schnitzler, Ho, Huang, Boudin, Sugawara, and Aizawa}]{schnitzler2024morehopqa}
Julian Schnitzler, Xanh Ho, Jiahao Huang, Florian Boudin, Saku Sugawara, and Akiko Aizawa. 2024.
\newblock Morehopqa: More than multi-hop reasoning.
\newblock \emph{arXiv preprint arXiv:2406.13397}.

\bibitem[{Shi et~al.(2024)Shi, You, Huang, Li, and Xiong}]{shi2024corecode}
Dan Shi, Chaobin You, Jiantao Huang, Taihao Li, and Deyi Xiong. 2024.
\newblock Corecode: A common sense annotated dialogue dataset with benchmark tasks for chinese large language models.
\newblock In \emph{Proceedings of the AAAI Conference on Artificial Intelligence}, volume~38, pages 18952--18960.

\bibitem[{Sun et~al.(2024)Sun, Huang, Wu, Gu, Li, Zhang, Yan, and He}]{sun2024benchmarking}
Jiaxing Sun, Weiquan Huang, Jiang Wu, Chenya Gu, Wei Li, Songyang Zhang, Hang Yan, and Conghui He. 2024.
\newblock Benchmarking chinese commonsense reasoning of llms: from chinese-specifics to reasoning-memorization correlations.
\newblock \emph{arXiv preprint arXiv:2403.14112}.

\bibitem[{Sun et~al.(2019)Sun, Yu, Yu, and Cardie}]{sun2019investigatingpriorknowledgechallenging}
Kai Sun, Dian Yu, Dong Yu, and Claire Cardie. 2019.
\newblock \href {https://arxiv.org/abs/1904.09679} {Investigating prior knowledge for challenging chinese machine reading comprehension}.
\newblock \emph{Preprint}, arXiv:1904.09679.

\bibitem[{Tang and Yang(2024)}]{tang2024multihop}
Yixuan Tang and Yi~Yang. 2024.
\newblock Multihop-rag: Benchmarking retrieval-augmented generation for multi-hop queries.
\newblock \emph{arXiv preprint arXiv:2401.15391}.

\bibitem[{Team et~al.(2025)Team, Du, Gao, Xing, Jiang, Chen, Li, Xiao, Du, Liao et~al.}]{team2025kimi}
Kimi Team, Angang Du, Bofei Gao, Bowei Xing, Changjiu Jiang, Cheng Chen, Cheng Li, Chenjun Xiao, Chenzhuang Du, Chonghua Liao, et~al. 2025.
\newblock Kimi k1. 5: Scaling reinforcement learning with llms.
\newblock \emph{arXiv preprint arXiv:2501.12599}.

\bibitem[{Team(2024)}]{qwen2.5}
Qwen Team. 2024.
\newblock \href {https://qwenlm.github.io/blog/qwen2.5/} {Qwen2.5: A party of foundation models}.

\bibitem[{Team(2025)}]{team2025qwq}
Qwen Team. 2025.
\newblock Qwq-32b: Embracing the power of reinforcement learning.
\newblock \emph{URL: https://qwenlm. github. io/blog/qwq-32b}.

\bibitem[{Trivedi et~al.(2022)Trivedi, Balasubramanian, Khot, and Sabharwal}]{trivedi2022interleavingretrieval}
Harsh Trivedi, Niranjan Balasubramanian, Tushar Khot, and Ashish Sabharwal. 2022.
\newblock Interleaving retrieval with chain-of-thought reasoning for knowledge-intensive multi-step questions.
\newblock \emph{arXiv preprint arXiv: Arxiv-2212.10509}.

\bibitem[{Veuthey et~al.(2025)Veuthey, Majid, Hariharan, and Haimes}]{veuthey2025meqametaevaluationframeworkquestion}
Jaime~Raldua Veuthey, Zainab~Ali Majid, Suhas Hariharan, and Jacob Haimes. 2025.
\newblock \href {https://arxiv.org/abs/2504.14039} {Meqa: A meta-evaluation framework for question answer llm benchmarks}.
\newblock \emph{Preprint}, arXiv:2504.14039.

\bibitem[{Wake et~al.(2025)Wake, Chen, Lv, Li, Huang, Cai, Zheng, Cooper, Zhou, Hu, Zhang, Wang, Ji, Qiu, Zhu, Tian, Su, Zhang, Li, Song, Li, Liu, Hu, Wang, Zhou, Yang, Li, Zhu, Xie, Huang, He, Chen, Hu, Ren, Niu, Li, Zhao, Luo, Xu, Sha, Yan, Liu, Zhang, and Dai}]{wake2025yilightningtechnicalreport}
Alan Wake, Bei Chen, C.~X. Lv, Chao Li, Chengen Huang, Chenglin Cai, Chujie Zheng, Daniel Cooper, Fan Zhou, Feng Hu, Ge~Zhang, Guoyin Wang, Heng Ji, Howard Qiu, Jiangcheng Zhu, Jun Tian, Katherine Su, Lihuan Zhang, Liying Li, Ming Song, Mou Li, Peng Liu, Qicheng Hu, Shawn Wang, Shijun Zhou, Shiming Yang, Shiyong Li, Tianhang Zhu, Wen Xie, Wenhao Huang, Xiang He, Xiaobo Chen, Xiaohui Hu, Xiaoyi Ren, Xinyao Niu, Yanpeng Li, Yongke Zhao, Yongzhen Luo, Yuchi Xu, Yuxuan Sha, Zhaodong Yan, Zhiyuan Liu, Zirui Zhang, and Zonghong Dai. 2025.
\newblock \href {https://arxiv.org/abs/2412.01253} {Yi-lightning technical report}.
\newblock \emph{Preprint}, arXiv:2412.01253.

\bibitem[{wavewangyue()}]{NLPCC-MH}
wavewangyue.
\newblock Nlpcc-mh.
\newblock \url{https://github.com/wavewangyue/NLPCC-MH/tree/master}.

\bibitem[{Welbl et~al.(2018)Welbl, Stenetorp, and Riedel}]{welbl2018constructing}
Johannes Welbl, Pontus Stenetorp, and Sebastian Riedel. 2018.
\newblock Constructing datasets for multi-hop reading comprehension across documents.
\newblock \emph{Transactions of the Association for Computational Linguistics}, 6:287--302.

\bibitem[{Wu et~al.(2024)Wu, Yang, Wang, Okumura, and Zhang}]{wu2024cofca}
Jian Wu, Linyi Yang, Zhen Wang, Manabu Okumura, and Yue Zhang. 2024.
\newblock Cofca: A step-wise counterfactual multi-hop qa benchmark.
\newblock \emph{arXiv preprint arXiv:2402.11924}.

\bibitem[{Xu et~al.(2020)Xu, Hu, Zhang, Li, Cao, Li, Xu, Sun, Yu, Yu et~al.}]{xu2020clue}
Liang Xu, Hai Hu, Xuanwei Zhang, Lu~Li, Chenjie Cao, Yudong Li, Yechen Xu, Kai Sun, Dian Yu, Cong Yu, et~al. 2020.
\newblock Clue: A chinese language understanding evaluation benchmark.
\newblock \emph{arXiv preprint arXiv:2004.05986}.

\bibitem[{Yang et~al.(2018)Yang, Qi, Zhang, Bengio, Cohen, Salakhutdinov, and Manning}]{yang2018hotpotqa}
Zhilin Yang, Peng Qi, Saizheng Zhang, Yoshua Bengio, William Cohen, Ruslan Salakhutdinov, and Christopher~D Manning. 2018.
\newblock Hotpotqa: A dataset for diverse, explainable multi-hop question answering.
\newblock In \emph{Proceedings of the 2018 Conference on Empirical Methods in Natural Language Processing}, pages 2369--2380.

\bibitem[{Zhong et~al.(2023{\natexlab{a}})Zhong, Guo, Tu, Zhang, Liu, and Sun}]{zhong2023legal}
Haoxi Zhong, Yichao Guo, Cunchao Tu, Tianyang Zhang, Zhiyuan Liu, and Maosong Sun. 2023{\natexlab{a}}.
\newblock Legal judgment prediction via legal reading comprehension.
\newblock \emph{arXiv preprint arXiv:2303.11495}.

\bibitem[{Zhong et~al.(2023{\natexlab{b}})Zhong, Wu, Manning, Potts, and Chen}]{zhong-etal-2023-mquake}
Zexuan Zhong, Zhengxuan Wu, Christopher Manning, Christopher Potts, and Danqi Chen. 2023{\natexlab{b}}.
\newblock \href {https://doi.org/10.18653/v1/2023.emnlp-main.971} {{MQ}u{AKE}: Assessing knowledge editing in language models via multi-hop questions}.
\newblock In \emph{Proceedings of the 2023 Conference on Empirical Methods in Natural Language Processing}, pages 15686--15702, Singapore. Association for Computational Linguistics.

\bibitem[{Zhou et~al.(2025)Zhou, Gong, Bao, Qiu, Li, Zhu, Zhang, and Zhang}]{zhou2025reasoning}
Qiji Zhou, Yifan Gong, Guangsheng Bao, Hongjie Qiu, Jinqiang Li, Xiangrong Zhu, Huajian Zhang, and Yue Zhang. 2025.
\newblock Reasoning is all you need for video generalization: A counterfactual benchmark with sub-question evaluation.
\newblock \emph{arXiv preprint arXiv:2503.10691}.

\bibitem[{Zhu et~al.(2024)Zhu, Hwang, Dugan, and Callison-Burch}]{zhu2024fanoutqa}
Andrew Zhu, Alyssa Hwang, Liam Dugan, and Chris Callison-Burch. 2024.
\newblock Fanoutqa: A multi-hop, multi-document question answering benchmark for large language models.
\newblock \emph{arXiv preprint arXiv:2402.14116}.

\end{thebibliography}

\appendix

\section{Prompt Templates}
\label{sec:prompt}
We provide prompt templates for seed question and sub-question classification, sub-question filtering, multi-hop QA Composition, and multi-hop QA verification (Figures~\ref{fig:classify_question},~\ref{fig:verify_qa_quality},~\ref{fig:multi_hop_qa_prompt}, and~\ref{fig:verify_multi_hop_qa}). The entire pipeline is executed using round-robin and cross-model verification strategies with several state-of-the-art LLMs, including GPT-4o, DeepSeek-R1, and Doubao-1.5-Pro.

\begin{figure*}[htbp]
\begin{tcolorbox}
\begin{CJK}{UTF8}{gbsn}
你是一名专业的数据标注员，现在需要你对给定的问题进行分类。我们将问题按照其知识领域划分为以下六个主类目：\\
1. \textbf{中国文化}：涵盖中国传统文化、历史人物、节日习俗、国粹等相关内容。\\
2. \textbf{人文社科}：包括哲学、文学、历史、语言学、教育学、社会学、心理学等学科内容。\\
3. \textbf{工程与技术}：涉及计算机、人工智能、通信、电子、机械、建筑等技术和工程领域。\\
4. \textbf{生活与艺术}：包含饮食健康、日常生活、体育、娱乐、电影、音乐、美术等内容。\\
5. \textbf{社会}：关注政治、经济、法律、军事、传媒、国际关系等社会制度与结构问题。\\
6. \textbf{自然科学}：包括数学、物理、化学、生物、天文、地理等自然科学相关问题。\\

你的任务是为每一个问题指定最相关的一级类目。请确保分类准确、不交叉、不模糊，避免主观解释，依据知识主题进行客观判断。\\

\#\#\# 以下是一些示例 \#\#\#\\
问题：屈原是哪一历史时期的诗人？\\
分类：中国文化\\

问题：中国第一颗人造卫星的名称是什么？\\
分类：工程与技术\\

问题：DNA分子的双螺旋结构是由谁发现的？\\
分类：自然科学\\

请你根据上述规范，对接下来的子问题进行分类。

\end{CJK}
\end{tcolorbox}
\caption{The prompt for reclassifying seed factual questions into six domains.}
\label{fig:classify_question}
\end{figure*}

\begin{figure*}[htbp]
\begin{tcolorbox}
\begin{CJK}{UTF8}{gbsn}

你是一名专业的数据标注员，负责对给定的问答对进行全面评估，并判断其是否符合高质量构建标准。请严格按照以下四项评估标准执行：

1. \textbf{问题结构合理、语义清晰}：问题应具备完整的结构和清晰的逻辑，语义明确，避免出现模糊或歧义的表达，如指代不明（例如“他”、“那个地方”）或缺少必要上下文。问题必须具有可回答性，其答案应为明确且可验证的客观事实。

2. \textbf{问题内容具备事实性与时间稳定性}：问题应聚焦于客观事实，避免包含主观判断或观点性内容（如“你如何看待……”），同时应排除时间敏感型问题（如“现任领导是谁”、“今年诺贝尔奖得主是谁”等随时间变化的问题）。

3. \textbf{答案准确、可验证}：答案应符合当前主流知识，不得包含过时、错误或未经验证的信息。你可以结合常识及相关文档内容判断答案的准确性。若答案存在疑问，请标记为“不合格”并简要说明原因。

4. \textbf{答案为明确、可数且独立的事实单元}：答案应为具体且清晰可识别的实体，如人物姓名、地点、机构名称或具体时间等，不应为笼统描述或抽象概念（如“很多国家”、“西方文化”等）。若存在多个独立且可数的正确答案，请用“/”分隔列出所有合理选项。

请依据以上标准，对每组问答对逐项审核，判断其是否“合格”或“不合格”，并附简要说明。\\

\#\#\# 以下是一些示例 \#\#\#\\
\textbf{问题}：新中国是在哪什么时候成立的？\\
\textbf{答案}：1949年\\
\textbf{评价}：问题表达不够明确，无法判定需回答具体日期还是年份。\\
\textbf{判断}：【不合格】\\

\vspace{0.5em}
\textbf{问题}：你认为日本最好的城市是哪一个？\\
\textbf{答案}：东京\\
\textbf{评价}：问题属于主观性范畴，缺乏客观可验证的标准。\\
\textbf{判断}：【不合格】\\

\vspace{0.5em}
\textbf{问题}：现任美国总统是谁？\\
\textbf{答案}：唐纳德·特朗普\\
\textbf{评价}：问题时效性强，答案会随时间变化。\\
\textbf{判断}：【不合格】

\vspace{0.5em}
\textbf{问题}：河北省的气候条件是什么，有什么特点？\\
\textbf{答案}：河北省属温带大陆性季风气候，具有春季干燥...\\
\textbf{评价}：问题指示不够具体，答案过于冗长且非单一实体，难以归类为事实单元。\\
\textbf{判断}：【不合格】

\vspace{1em}
请参照上述标准和示例，以相同格式对接下来的问答对进行评估。

\end{CJK}
\end{tcolorbox}
\caption{The prompt for sub-question quality verification.}
\label{fig:verify_qa_quality}
\end{figure*}

\begin{figure*}[htbp]
\begin{tcolorbox}
\begin{CJK}{UTF8}{gbsn}

你是一名专业的数据标注员，负责将多个单跳问答（即若干事实性子问题及其答案）整合为自然、连贯、逻辑清晰的多跳问答数据。请严格按照以下要求完成任务：

\vspace{1em}
\textbf{任务目标}：

根据提供的若干子问题及其对应答案，构造一个多跳问题，使其必须依赖多个中间推理步骤才能得出最终答案。合成后的问题应具备自然语言表达能力、逻辑连贯性，并指向一个唯一、具体、可验证的事实答案。

\vspace{1em}
\textbf{构造要求}：

1. \textbf{唯一且具体的最终答案}：  
最终问题应指向一个明确、可验证的事实性答案，不能存在歧义或多种可能性。中间子问题的答案应共同服务于该最终答案的推出。答案必须为具体、可识别的实体，如人物姓名、地名、机构名称或确切时间等，避免使用模糊描述或抽象概念。

2. \textbf{顺序推理、逻辑通顺}：  
多跳问题应展现清晰、线性推进的推理链，各步骤之间应通过因果、指代或其他自然语言衔接方式建立联系。中间答案不得直接出现在问题中，须通过代词、称谓或上下文信息进行指代。不得出现信息冗余、逻辑重复或循环依赖的问题结构。

3. \textbf{语言自然、表达流畅}：  
合成的问题应具备完整、清晰、符合语言习惯的自然语言表达。可根据需要添加适当的上下文信息以增强问题的可读性和连贯性，但不得显式暴露中间答案内容。

4. \textbf{内容安全、符合伦理}：  
所构造的问题不得涉及敏感、争议、歧视、暴力等内容，须符合基本伦理和安全规范。

\vspace{1em}
\#\#\# 以下是一些示例 \#\#\#\\

\textbf{子问题：}\\
\texttt{['1954年获得诺贝尔化学奖的是谁？',\\
'莱纳斯·卡尔·鲍林曾在哪所大学取得博士学位？',\\
'加州理工学院创办于哪一年？']}

\textbf{答案：}\\
\texttt{['莱纳斯·卡尔·鲍林', '加州理工学院', '1891年']}

\vspace{0.5em}
\textbf{合成的多跳问题：}1954年诺贝尔化学奖得主获得博士学位的大学创办于哪一年？\\
\textbf{最终答案：}1891年

\vspace{1em}
请你根据上述规范，对接下来的子问题集合进行合理整合，生成符合标准的多跳问题及其答案。

\end{CJK}
\end{tcolorbox}
\caption{Prompt for multi-hop QA composition from sub-questions.}
\label{fig:multi_hop_qa_prompt}
\end{figure*}

\begin{figure*}[htbp]
\begin{tcolorbox}
\begin{CJK}{UTF8}{gbsn}

你是一名专业的数据标注员，负责对已构造的多跳问答数据进行系统审核，判断其是否符合高质量构建标准。请严格依据以下评估维度，对每条多跳问答进行逐项检查，并将其标注为“合格”或“不合格”，必要时请简要说明不合格原因。

\vspace{1em}
\textbf{评估维度}：

1. \textbf{全局答案唯一性}：  
多跳问题必须导向一个明确、具体、可验证的最终唯一答案。不得存在语义模糊或多个合理选项。最终答案应为具象事实性实体，如人名、地名、机构或具体时间等。

2. \textbf{推理链条合理性}：  
问题应体现清晰的多跳推理路径，子问题之间具有明确的逻辑依赖关系。不得存在跳步（可省略中间步骤）、信息冗余、逻辑回环或中间答案显式暴露等问题。

3. \textbf{语言自然与指代衔接}：  
问题表达应符合自然语言使用习惯，语言流畅，语境连贯。中间实体须通过恰当的代词、称谓或上下文表达进行隐式衔接，避免直接重复出现。

4. \textbf{内容安全性}：  
问题内容不得涉及敏感、争议、歧视、暴力等信息，必须符合伦理规范和公共安全标准。

\vspace{1em}
\#\#\# 以下是一些示例 \#\#\#\\

\vspace{0.5em}
\textbf{多跳问题：}1954年诺贝尔化学奖得主获得博士学位的大学创办于哪一年？\\
\textbf{最终答案：}1891年\\
\textbf{评价：}推理链条完整，指代衔接自然，语言表达清晰，最终答案明确且可验证。\\
\textbf{判断：}【合格】

\vspace{0.5em}
\textbf{多跳问题：}1954年诺贝尔奖得主毕业的大学创办于什么时候？\\
\textbf{最终答案：}1891年\\
\textbf{评价：}问题表达模糊，未指定“诺贝尔什么奖”，毕业阶段不清，时间问法不明确。\\
\textbf{判断：}【不合格】

\vspace{0.5em}
\textbf{多跳问题：}1954年诺贝尔化学奖得主获得博士学位的大学的哪位校友在1954年获得了诺贝尔化学奖？\\
\textbf{最终答案：}莱纳斯·卡尔·鲍林\\
\textbf{评价：}存在逻辑回绕，推理链形成闭环；同时可通过最后子句直接获取答案，推理路径冗余。\\
\textbf{判断：}【不合格】

\vspace{1em}
请你严格依据上述评估标准，对多跳问答数据进行质量审核，并判断是否合格。

\end{CJK}
\end{tcolorbox}
\caption{Prompt for verifying the quality of composed multi-hop questions.}
\label{fig:verify_multi_hop_qa}
\end{figure*}

\begin{table*}[h]
\centering
\small
\setlength{\arrayrulewidth}{1pt}
\setlength{\tabcolsep}{12pt}

\renewcommand{\arraystretch}{1.8}
\begin{tabular}{|p{4cm}|p{2.5cm}|p{4cm}|p{1.5cm}|}
\hline
\textbf{Sub-Questions} & \textbf{Sub-Question Answers} & \textbf{Multihop Question} & \textbf{Final Answer} \\

\hline
\quad\newline
1.《少湖文集》的作者是谁? \newline
2.《少湖文集》的作者徐阶所处的朝代是哪个朝代？\newline
3. 明朝最终灭亡于哪一年？\newline
&
\quad\newline
徐阶\newline
明朝\newline\newline
1644年\newline
&
\quad\newline
《少湖文集》的作者所属某朝代, 该朝代最终灭亡于哪一年？
&
\quad\newline
1644年 \\

\hline
\quad\newline
1. 致力于方法和理论研究，新印象主义的倡导者是谁？\newline
2. 新印象主义画家乔治·修拉的哪个国家的公民？\newline
3. 法国国歌是什么？\newline
&
\quad\newline
乔治·修拉\newline
\newline法国\newline\newline
马赛曲
&
\quad\newline
致力于方法和理论研究，新印象主义的倡导者所属国家的国歌是什么？
&
\quad\newline
马赛曲 \\
\hline
\quad\newline
1. 联合国成立于哪一年？\newline
2. 1945年的诺贝尔文学奖获得者是谁？\newline
3. 诺贝尔文学奖得主加夫列拉·米斯特拉尔的国籍是哪个国家？\newline
&
\quad\newline
1945年\newline\newline
加夫列拉·米斯特拉尔\newline\newline
智利\newline
&
\quad\newline
联合国成立的同一年的诺贝尔文学奖的获得者的国籍是哪个国家？
&
\quad\newline
智利 \\
\hline

\quad\newline
1. 瑞典荷尔斯泰因-戈托普王朝的第二位国王是谁?\newline
2. 瑞典国王古斯塔夫三世去世于哪一年?\newline
3. 1792年在任的美国总统是谁?\newline
4. 美国总统乔治·华盛顿的出生地属于哪个州?\newline
5. 弗吉尼亚州的首府是哪个城市?\newline
6. 弗吉尼亚州里士满市的电话区号是什么？\newline 
&
\quad\newline
古斯塔夫三世 \newline\newline
1792年\newline\newline
乔治·华盛顿\newline\newline
弗吉尼亚州\newline\newline
里士满市\newline\newline
804\newline\newline
&
\quad\newline
瑞典荷尔斯泰因-戈托普王朝的第二位国王，他去世那一年在任的美国总统，该总统的出生地属于某个州，这个州的首府的电话区号是什么？
&
\quad\newline
804 \\
\hline
\quad\newline
1. 中国香港特别行政区第五任行政长官是谁？\newline
2. 香港第五任行政长官林郑月娥的配偶是谁？\newline
3. 林郑月娥的配偶林兆波取得了哪所大学的博士学位？\newline
4. 担任剑桥大学教师且在2017年获得诺贝尔化学奖的人是谁？\newline
5. 2017年诺贝尔化学奖获得者理查德·亨德森获得了哪所大学的物理学学士学位？\newline
6. 2002年担任英国爱丁堡大学校长的人是谁？\newline
&
\quad\newline
林郑月娥\newline\newline
林兆波\newline\newline
剑桥大学\newline\newline
理查德·亨德森\newline\newline
爱丁堡大学\newline\newline
蒂莫西·奥谢\newline
&
\quad\newline
中国香港特别行政区第五任行政长官的配偶取得了某所大学的博士学位，担任该大学教授且在2017年获得诺贝尔化学奖的人获得了一所大学的物理学学士学位，2002年担任该大学校长的人是谁？
&
\quad\newline
蒂莫西·奥谢 \\
\bottomrule
\end{tabular}
\caption{Data examples of three-hops and six-hops multihop questions.}
\label{tab:data_exmaple}
\end{table*}

\section{Data Examples}
Table~\ref{tab:data_exmaple} presents some constructed examples of three-hops and six-hops multihop questions.

\section{Case Study}
\label{sec:appendix_case_model}
In this section, we present case studies of model responses, including examples of three-hop reasoning in Table~\ref{Table:3-hops_case}, six-hop reasoning in Table~\ref{Table:6-hops_case}, and cases incorporating RAG in Table~\ref{Table:rag_case}.

\begin{table*}[htbp]
\centering
\small
\setlength{\arrayrulewidth}{1pt}
\setlength{\tabcolsep}{12pt}

\renewcommand{\arraystretch}{1.8}
\begin{tabularx}{\textwidth}{|l|X|}
\hline
\textbf{Question} & 
全世界最长山脉的最高峰位于某个国家境内, 该国家的首都是哪座城市？ \\
\hline
\textbf{GPT-4o} & 
全世界最长的山脉是安第斯山脉，其最高峰是阿空加瓜山（Aconcagua），位于阿根廷境内。阿根廷的首都是布宜诺斯艾利斯。

<答案>布宜诺斯艾利斯</答案>\\

\hline
\textbf{Gemini-2.5-flash} & 
全世界最长的山脉是安第斯山脉。
安第斯山脉的最高峰是阿空加瓜山（Aconcagua）。
阿空加瓜山位于阿根廷境内。
阿根廷的首都城市是布宜诺斯艾利斯。

<答案>布宜诺斯艾利斯</答案>\\
\hline
\textbf{GLM-4-air} & 
全世界最长山脉是安第斯山脉，其最高峰是阿空加瓜山，位于阿根廷境内。阿根廷的首都是布宜诺斯艾利斯。

<答案>布宜诺斯艾利斯</答案>>\\
\hline

\textbf{DeepSeek-R1} & 
嗯，这个问题看起来有点挑战性。首先，题目是说全世界最长的山脉的最高峰位于某个国家境内，然后问这个国家的首都是哪座城市。我需要一步步来拆解这个问题。

首先，我得确定全世界最长的山脉是什么。根据以前学过的地理知识，最长的山脉应该是安第斯山脉吧？安第斯山脉位于南美洲，沿着西海岸延伸，经过多个国家，比如哥伦比亚、厄瓜多尔、秘鲁、玻利维亚、智利、阿根廷等等。安第斯山脉确实是世界上最长的山脉，长度超过7000公里，对吧？

接下来，问题提到的是这个最长山脉的最高峰。那安第斯山脉的最高峰是哪一座呢？我记得安第斯山脉的最高峰是阿空加瓜山（Aconcagua），位于阿根廷境内。阿空加瓜山的海拔大约是6961米，是南美洲的最高峰，也是亚洲之外最高的山峰。所以，最高峰阿空加瓜山应该是在阿根廷的境内。

那阿根廷的首都是哪里呢？阿根廷的首都是布宜诺斯艾利斯（Buenos Aires）。不过，我得确认一下这些信息是否正确，避免记错了。比如，有没有可能安第斯山脉的最高峰在别的国家？比如智利或者秘鲁？比如，查一下阿空加瓜山的位置，是否真的属于阿根廷。根据资料，阿空加瓜山位于阿根廷门多萨省，靠近智利边境，但主权属于阿根廷。所以确认是阿根廷。

那总结一下，最长山脉是安第斯山脉，最高峰阿空加瓜山在阿根廷，阿根廷的首都是布宜诺斯艾利斯。所以答案应该是布宜诺斯艾利斯。

<答案>布宜诺斯艾利斯</答案> \\
\hline
\textbf{Gemini-2.5-Pro} & 
<答案>布宜诺斯艾利斯</答案>
**推理过程：**
1.  **全世界最长的山脉**：安第斯山脉 (Andes Mountains)。
2.  **安第斯山脉的最高峰**：阿空加瓜山 (Mount Aconcagua)，海拔6960.8米。
3.  **阿空加瓜山所在的国家**：阿根廷 (Argentina)。
4.  **阿根廷的首都**：布宜诺斯艾利斯 (Buenos Aires)。 \\
\hline
\textbf{GLM-z1-air} &
嗯，这个问题看起来是要找出全世界最长山脉的最高峰所在国家的首都。首先，我需要确定哪座山脉是世界上最长的。我记得安第斯山脉是世界上最长的山脉，沿着南美洲的西海岸延伸，从哥伦比亚一直延伸到智利和阿根廷，对吧？那它的最高峰应该是阿空加瓜山，对吧？阿空加瓜山位于阿根廷境内，所以最高峰所在的国家是阿根廷。接下来，阿根廷的首都是布宜诺斯艾利斯。不过，我得再确认一下这些信息是否正确，避免记错了。

首先，确认最长山脉。安第斯山脉确实是世界最长的山脉，长度超过7000公里，而喜马拉雅山脉虽然高，但长度较短。所以最高峰是安第斯山脉的阿空加瓜山，海拔6961米，位于阿根廷。然后阿根廷的首都是布宜诺斯艾利斯，没错。所以答案应该是布宜诺斯艾利斯。
<答案>布宜诺斯艾利斯</答案> \\
\bottomrule
\end{tabularx}
\caption{Comparison of responses from different LLMs on three-hops quesiton.}
\label{Table:3-hops_case}
\end{table*}

\begin{table*}[htbp]
\centering
\small
\setlength{\arrayrulewidth}{1pt}
\setlength{\tabcolsep}{12pt}

\renewcommand{\arraystretch}{1.8}
\begin{tabularx}{\textwidth}{|l|X|}
\hline
\textbf{Question} & 
瑞士面积最大州的首府位于某条河流右岸，这条河流最终在某城市附近注入北海，2009年当选该市市长的人出生于哪个国家？？ \\
\hline
\textbf{GPT-4o} & 
瑞士面积最大的州是格劳宾登州（Graubünden），其首府是库尔（Chur）。库尔位于莱茵河右岸。莱茵河最终在荷兰鹿特丹附近注入北海。2009年当选鹿特丹市市长的是艾哈迈德·阿布塔莱布（Ahmed Aboutaleb），他出生于摩洛哥。

<答案>摩洛哥</答案>\\

\hline
\textbf{Gemini-2.5-flash} & 
瑞士面积最大州（格劳宾登州/Graubünden）的首府是库尔（Chur）。
库尔位于莱茵河（Rhine）的右岸（具体来说是阿尔卑斯莱茵河段的右岸）。
莱茵河最终在荷兰的鹿特丹（Rotterdam）附近注入北海。
2009年当选鹿特丹市长的人是艾哈迈德·阿布塔勒布（Ahmed Aboutaleb），他于2009年1月5日上任。
艾哈迈德·阿布塔勒布出生于摩洛哥。

<答案>摩洛哥</答案>\\
\hline
\textbf{GLM-4-air} & 
要解决这个问题，我们需要逐步分析题目中的信息：

1. **瑞士面积最大州的首府**：瑞士面积最大的州是格劳宾登州（Graubünden），其首府是库尔（Chur）。

2. **库尔位于某条河流右岸**：库尔位于莱茵河（Rhine River）的右岸。

3. **莱茵河最终在某城市附近注入北海**：莱茵河最终在鹿特丹（Rotterdam）附近注入北海。

4. **2009年当选鹿特丹市长的人**：2009年当选鹿特丹市长的是伊沃·范德拉格（Ivo Opstelten），他出生于荷兰。

因此，2009年当选鹿特丹市长的人出生于荷兰。

<答案>荷兰</答案>\\
\hline

\textbf{DeepSeek-R1} & 
嗯，这个问题看起来有点复杂，不过让我慢慢理清楚。首先，题目问的是瑞士面积最大的州的首府位于某条河流的右岸，这条河流最终在某个城市附近注入北海。然后，2009年当选该市市长的人出生在哪个国家。我需要一步步拆解这些信息。

首先，瑞士面积最大的州是哪个呢？瑞士的州有很多，比如苏黎世州、伯尔尼州、格劳宾登州等等。我记得瑞士面积最大的州应该是格劳宾登州（Graubünden），因为那里有很多山区，面积比较大。不过不确定的话，可能需要确认一下。查资料的话，瑞士面积最大的州确实是格劳宾登州，首府是库尔（Chur）。

接下来，库尔位于哪条河流的右岸呢？库尔位于莱茵河的一条支流旁边，可能是前莱茵河（Vorderrhein）或者后莱茵河（Hinterrhein），不过更可能的是，整个莱茵河的上游部分。不过莱茵河的干流确实流经库尔附近，库尔坐落在莱茵河的右岸吗？需要确认。莱茵河在瑞士境内分为前莱茵河和后莱茵河，汇合后成为阿尔卑斯莱茵河，流经库尔。库尔位于阿尔卑斯莱茵河的右岸，对吗？是的，库尔确实在莱茵河的右岸。

然后，这条河流最终注入北海，而注入北海的城市是哪里呢？莱茵河最终在荷兰的鹿特丹附近流入北海。所以问题中的“某城市”应该是指鹿特丹。那么，2009年当选鹿特丹市长的人是谁呢？需要查一下鹿特丹的市长在2009年的选举情况。当时的市长是艾哈迈德·阿布塔莱布（Ahmed Aboutaleb），他于2009年1月就任鹿特丹市长。他的出生国家是摩洛哥，因为他出生于摩洛哥的贝尼·斯迪尔（Beni Sidel）。

所以，整个链条应该是：瑞士最大的州是格劳宾登，首府库尔位于莱茵河右岸，莱茵河最终在鹿特丹附近入北海，2009年当选鹿特丹市长的是艾哈迈德·阿布塔莱布，出生在摩洛哥。

<答案>摩洛哥</答案>\\
\hline
\textbf{Gemini-2.5-Pro} & 
逐步分析： 1.  **瑞士面积最大州的首府**：    *   瑞士面积最大的州是格劳宾登州 (Graubünden)。    *   格劳宾登州的首府是库尔 (Chur)。 2.  **位于某条河流右岸**：    *   库尔 (Chur) 位于普莱苏尔河 (Plessur) 的右岸，普莱苏尔河在库尔附近汇入莱茵河 (Rhine)。库尔本身也通常被描述为坐落在莱茵河谷。考虑到后续问题，这里指的“某条河流”更可能是指莱茵河。库尔确实在莱茵河（阿尔卑斯莱茵河段）的右岸（东岸）。 3.  **这条河流最终在某城市附近注入北海**：    *   莱茵河 (Rhine) 最终流经荷兰，在鹿特丹 (Rotterdam) 附近通过其三角洲注入北海。 4.  **2009年当选该市市长的人出生于哪个国家？**    *   该市指的是鹿特丹 (Rotterdam)。    *   2009年当选（实际是2009年1月5日上任）鹿特丹市长的是艾哈迈德·阿布塔莱卜 (Ahmed Aboutaleb)。    *   艾哈迈德·阿布塔莱卜出生于摩洛哥 (Morocco)。 综上所述，最终答案是摩洛哥。<答案>摩洛哥</答案> \\

\bottomrule
\end{tabularx}
\caption{Comparison of responses from different LLMs on six-hops quesiton.}
\label{Table:6-hops_case}
\end{table*}

\begin{table*}[htbp]
\centering
\small
\setlength{\arrayrulewidth}{1pt}
\setlength{\tabcolsep}{12pt}

\renewcommand{\arraystretch}{1.8}
\begin{tabularx}{\textwidth}{|l|X|}
\hline
\textbf{Question} & 
漫画《寻找满月》的原作者在2011年与某月刊解除专属契约，该月刊创刊的那一年5月14日成立了一个组织，该组织起草人的长子是谁？ \\
\hline
\textbf{Doubao-1.5-pro} & 
\#\#\# 步骤一：确定漫画《寻找满月》原作者及相关月刊
漫画《寻找满月》的原作者是种村有菜。2011年种村有菜与集英社《RIBON》月刊解除专属契约。

\#\#\# 步骤二：明确《RIBON》创刊时间及对应组织
《RIBON》创刊于1955年，1955年5月14日成立的组织是华沙条约组织（简称华约）。

\#\#\# 步骤三：找出华约起草人及长子信息
华约的起草人一般认为是苏联领导人赫鲁晓夫。赫鲁晓夫有两个儿子，长子是列昂尼德·赫鲁晓夫。列昂尼德在第二次世界大战中牺牲。

所以，该组织起草人的长子是列昂尼德·赫鲁晓夫。 \\
\hline
\textbf{Doubao-1.5-pro with RAG} & 
漫画《寻找满月》的原作者是种村有菜。2011年，她与日本集英社《RIBON》月刊解除专属契约。《RIBON》月刊创刊于1955年。1955年5月14日成立的组织是华沙条约组织。 华沙条约组织的建立是苏联等国为应对西方国家的威胁而采取的集体防御措施，是多个国家共同参与的结果，并没有明确单一的起草人，所以，根据现有信息，无法确定其起草人的长子是谁。 \\

\hline
\hline
\textbf{Question} & 
2018年11月担任中国社会科学院国际法研究所所长的人在1986年毕业于某所大学，2022年6月担任该大学校长的人在2020年出版了一本书，该书出版社的总部位于哪个城市？\\

\hline
\textbf{Doubao-1.5-pro} & 
仅根据你提供的这些信息，无法明确具体涉及的人物、大学以及书籍出版社等内容，也就无法确定该书出版社总部所在城市。你可以补充更多详细信息，比如相关人物姓名、大学名称等，以便我更准确地为你解答。  \\
\hline
\textbf{Doubao-1.5-pro with RAG} & 
2018年11月担任中国社会科学院国际法研究所所长的是莫纪宏。他1986年毕业于北京大学法律学系。2022年6月担任北京大学校长的是郝平。 在2020年，由北京大学中国政治学研究中心主持编撰、大百科全书出版社出版了《政治通鉴》，北京大学校长郝平出席了新书发布会。大百科全书出版社的总部位于北京，地址是北京市西城区阜成门北大街17号。\\

\bottomrule
\end{tabularx}
\caption{Examples of the effect of RAG on multi-hop quesitons.}
\label{Table:rag_case}
\end{table*}

\end{CJK}

\end{document}